\documentclass[journal]{IEEEtran}
\usepackage{amsmath,amsfonts}
\usepackage{algorithm}
\usepackage{algpseudocode}
\usepackage{array}
\usepackage[caption=false,font=normalsize,labelfont=sf,textfont=sf]{subfig}
\usepackage{textcomp}
\usepackage{stfloats}
\usepackage{url}
\usepackage{verbatim}
\usepackage{graphicx}
\usepackage{cite}
\usepackage{multirow}
\usepackage{tabularx}
\usepackage[table,xcdraw]{xcolor}
\usepackage{subfig}
\usepackage{extarrows}
\usepackage{booktabs}
\usepackage{array}
\usepackage{makecell}
\usepackage{adjustbox}
\usepackage{pdflscape}
\usepackage{xcolor}
\usepackage{color, colortbl}
\usepackage[colorlinks,linkcolor=blue,citecolor=blue]{hyperref}
\begin{document}

\title{UmambaTSF: A U-shaped Multi-Scale Long-Term Time Series Forecasting Method Using Mamba}

\author{Li Wu, Wenbin Pei, Jiulong Jiao, Qiang Zhang
        % <-this % stops a space

  \thanks{
    This work was supported by the National Key Research and Development Program of China under grant 2021ZD0112400, the National Natural Science Foundation of China under grants 42265010, 62162053, 62206041, 12371516 and U21A20491, and the NSFC-Liaoning Province United Foundation under grant U1908214, the 111 Project under grant D23006, the Liaoning Revitalization Talents Program under grant XLYC2008017, and China University Industry-University-Research Innovation Fund under grants 2022IT174, Natural Science Foundation of Liaoning Province under grant 2023-BSBA-030, and an Open Fund of National Engineering Laboratory for Big Data System Computing Technology under grant SZU-BDSC-OF2024-09.%(Corresponding author: Qiang Zhang and Wenbin Pei.)
    }

\thanks{Li Wu and Jiulong Jiao are with the School of Computer Science and Technology, Dalian University of Technology, Dalian, 116024, China; Qinghai University, Xining, 810016, China. (e-mail: wuli777@mail.dlut.edu.cn;jiaojiulong@mail.dlut.edu.cn)}% <-this % stops a space

 \thanks{Wenbin Pei and Qiang Zhang are with the School of Computer Science and Technology, Dalian University of Technology, Dalian 116024, China; Key Laboratory of Social Computing and Cognitive Intelligence (Dalian University of Technology), Ministry of Education, Dalian 116024, China. (e-mail: peiwenbin@dlut.edu.cn;zhangq@dlut.edu.cn)}  }

%\author{IEEE Publication Technology,~\IEEEmembership{Staff,~IEEE,}
        % <-this % stops a space
%\thanks{This paper was produced by the IEEE Publication Technology Group. They are in Piscataway, NJ.}% <-this % stops a space
%\thanks{Manuscript received April 19, 2021; revised August 16, 2021.}}


%\markboth{IEEE Transactions on Cybernetics }{ \MakeLowercase{\textit{Wu et al.}}: U-shaped Multi-Scale Long-Term Time Series Forecasting}

\markboth{IEEE Transactions on XXXXX,~Vol.~0, No.~0, December~0}{
\MakeLowercase{\textit{Wu et al.}}: U-shaped Multi-Scale Long-Term Time Series Forecasting}
\maketitle

\begin{abstract}
Multivariate Time series forecasting is crucial in domains such as transportation, meteorology, and finance, especially for predicting extreme weather events. State-of-the-art methods predominantly rely on Transformer architectures, which utilize attention mechanisms to capture temporal dependencies. However, these methods are hindered by quadratic time complexity, limiting the model's scalability with respect to input sequence length. This significantly restricts their practicality in the real world. Mamba, based on state space models (SSM), provides a solution with linear time complexity, increasing the potential for efficient forecasting of sequential data. In this study, we propose UmambaTSF, a novel long-term time series forecasting framework that integrates multi-scale feature extraction capabilities of U-shaped encoder-decoder multilayer perceptrons (MLP) with Mamba's long sequence representation. To improve performance and efficiency, the Mamba blocks introduced in the framework adopt a refined residual structure and adaptable design, enabling the capture of unique temporal signals and flexible channel processing. In the experiments, UmambaTSF achieves state-of-the-art performance and excellent generality on widely used benchmark datasets while maintaining linear time complexity and low memory consumption.
%In this study, we propose UmambaTSF, a novel method that combines U-Net's feature extraction capabilities with Mamba's long sequence representation. 
%we propose a new method named UmambaTSF, which integrates the feature extraction capability of U-Net with the long sequence representation of Mamba. 
\end{abstract}

\begin{IEEEkeywords}
Multivariate time series forecasting, Mamba, Multi-scale features, linear scalability.
\end{IEEEkeywords}

\section{Introduction}
\IEEEPARstart{T}{ime} series forecasting remains a fundamental problem in deep learning, with a wide range of real-world applications where future sequence behaviors are inferred from historical data. Models structured founded on convolutional neural networks (CNNs) and recurrent neural networks (RNNs) have traditionally addressed the complexities of temporal dependencies \cite{liu2022scinet,su2024mdcnet,salinas2020deepar}. However, the advent of transformer-based architectures has gained prominence due to their exceptional self-attention capabilities \cite{vaswani2017attention}. In Particular, the introduction of patch-based transformers, e.g., PatchTST \cite{nie2023a}, effectively captures both short-term and long-term temporal dependencies, establishing them as exemplary models in long-term forecasting. 

With growing demands in practical scenarios and advancements in deep learning techniques, there has been a concerted effort to extend the forecasting horizon of these models. Autofomer \cite{wu2021autoformer} automates periodic decomposition and aggregates similar subsequences to mitigate the entanglement and inefficiencies in self-attention mechanisms, thereby extending forecast durations. Subsequently, FEDformer \cite{zhou2022fedformer}, non-stationary Transformers \cite{liu2022non} and iTranformer \cite{liu2024itransformer} are exemplary models that continue the exploration of long forecasting horizons. However, transformer-based models have been typically criticized for their quadratic time complexity, despite a substantial body of work aimed at addressing this challenge \cite{zhou2021informer,liu2021pyraformer, pmlr-v162-wu22m}. The computation of the self-attention mechanism is limited to the receptive window, making it unable to directly understand elements outside this window. In general, a small window size results in poor model performance, while a large window size dramatically increases computational complexity. Meanwhile, linear models \cite{zeng2023transformers,li2023revisiting,wang2024lightweight}, due to their simple and efficient temporal representation and strong interpretability, have successfully demonstrated competitive prediction accuracy. Despite that, empirical comparisons reveal that linear models encounter challenges in complex temporal contexts and tasks requiring extended forecasting horizons. This is primarily due to their limited capacity for non-linear expression and insufficient historical context windows to effectively capture intricate sequence dependencies.

The Mamba model, based on a state-space model (SSM), has recently shown performance comparable to the Transformer in sequence data representation \cite{gu2023mamba,wang2024mamba,ahamed2024timemachine}. It also excels in graph-structured data \cite{behrouz2024graph}, image processing \cite{zhu2024vision,liu2024swin,patro2024simba}, and multimodal learning \cite{zhao2024cobra,qiao2024vl}, offering superior performance and computational efficiency. 
Mamba captures long-term correlations and features a context-aware selective mechanism with hidden attention, making it easier to infer long sequences with linear time complexity\cite{ali2024hidden}. This advancement creates new opportunities for efficient long-sequence forecasting. However, there are several open challenges in leveraging the Mamba model for time series forecasting. 

\textbf{Challenge 1}: The Mamba model is initially designed to handle long sequence inputs, where efficiently capturing both short-term and long-term temporal dependencies is challenging. Patch-based and sliding window methods disrupt the long cyclic signals crucial to the Mamba model. Additionally, when directly processing input sequences, the Mamba model also struggles to capture multi-scale temporal features, leading to suboptimal use of the information density in time series data. Since time series data often show varying cycles and trends \cite{angryk2020multivariate, wang2023micn, feng2023multi}, developing a Mamba model for multi-scale information extraction is essential.

%Since time series data often exhibits a variety of cycles and trends, developing a multi-scale information extraction Mamba model is imperative. 

\textbf{Challenge 2}: Time series data often contain overlapping signals, including cyclical, trend-based, seasonal, and stochastic \cite{yue2022ts2vec, dai2024periodicity}, making feature extraction at each scale challenging. A standalone Mamba model with a limited state space struggles to capture complex signal variations \cite{ahamed2024timemachine}. The major challenge is how a simple Mamba configuration can extract the unique information in each cyclic signal, balancing the accuracy of temporal feature representation while improving model efficiency.

\textbf{Challenge 3}: Most studies employ either channel independence or channel parallelism to process multivariate time series, but real-world variable interactions are often more complex. PatchTST \cite{nie2023a} and GPT4TS \cite{zhou2023one} adopt channel independence, processing each channel separately. Differently, iTransformer \cite{liu2024itransformer} utilizes channel parallelism, treating channels as multi-dimensional features. For strongly correlated datasets, channel parallelism may need refinement to better capture variable interactions. Therefore, a more flexible processing approach is essential to address diverse relationships.

This paper introduces an innovative framework, UmambaTSF, based on a combination of mamba and linear layers, for multivariate time series forecasting. To address Challenge 1, we introduce a U-shaped multi-scale feature extraction module in UmambaTSF, incorporating Mamba structures. This module leverages Mamba's capacity to capture temporal dependencies across multiple scales, extracting time-series features at each scale to fully exploit the input data. For Challenge 2, inspired by the N-Beats model \cite{Oreshkin2020N-BEATS:}, we propose residual mamba layers, which utilize residual information to iteratively remove noise and redundant signals, uncovering unique information at each scale. Furthermore, to balance training efficiency, the state expansion factor in Mamba is maintained as small as possible. To tackle Challenge 3, UmambaTSF features a flexible Channel-Adaptable Mamba module, consisting of a single Mamba block. This module adjusts to varying channel correlations, allowing for different processing methods that enhance the model's adaptability across diverse datasets. Our contributions are summarized as:

\begin{itemize}
    \item[1).] We design a multi-feature extractor that captures multiple periodic patterns from time series data across different scales, incorporating both long-term and short-term temporal dependencies. 
%We propose a method for extracting temporal context information from multi-dimensional time series, termed the Mamba Block, which consists of multiple foundational mamba components. This method aids the U-shaped multi-scale feature extractor in capturing the dependencies of temporal features at each scale. Additionally, it accommodates different interaction relationships across datasets, while satisfying channel independence, channel interdependence, and deep channel integration for multivariate information processing in various scenarios.
\item[2).] We develop residual Mamba layers to eliminate overlapping and noisy signals in time series data by leveraging residual information, enabling more accurate temporal feature representation at each scale. 

\item[3).] We propose a flexible Channel-Adaptable Mamba module that effectively adapts to complex channel relationships in multivariate time series, optimizing information transformation for forecasting across various scenarios.
\end{itemize}

\begin{figure}
\vspace{-1em}
\centering
\includegraphics[width=0.40\textwidth]{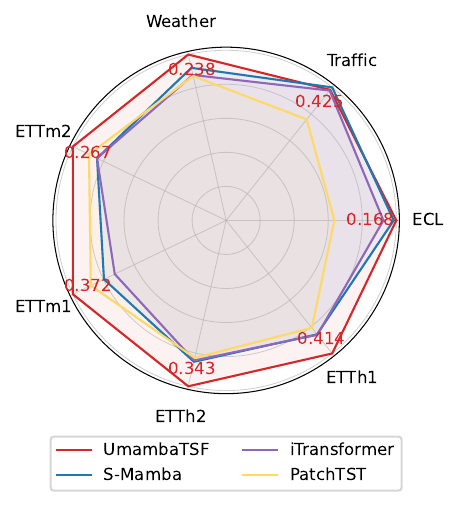}
\caption{Average performance (MSE) between UmambaTSF and the latest SOTA models on seven public real-world datasets. The center of the circle denotes the maximum error, while points nearer to the boundary indicate better performance.}
\label{fig:radar}
\end{figure}

According to the results of the experiments, UmambaTSF achieves state-of-the-art (SOTA) predictive accuracy (as shown in Fig. \ref{fig:radar}) with the complexity of $O(L)$ for length-$L$ series and minimal memory usage on public real-world datasets. We also conduct comprehensive analyses on the generalization ability of UmambaTSF, highlighting the potential of SSM-based methods in time series forecasting.

\section{Background}
In this section, we introduce the problem definition of multivariate time series forecasting, followed by the related works in Transformer-based Time Series Forecasting, MLP-based Time Series Forecasting and SSM-based sequence modeling.

\subsection{Multivariate Time Series Forecasting Problem Definition}
In multivariate time series forecasting task, historical sequence is defined as $X =\left[x_1, x_2, \ldots, x_L\right] \in \mathcal{R}^{N \times L}$, where $L$ represents the length of the historical data and $N$ indicates the number of variables. The ground truth future sequence of length $T$ is $Y =\left[x_{L+1}, x_{L+2}, \ldots, x_{L+T}\right] \in \mathcal{R}^{N \times T}$. Based on the sequential information and patterns in $X$, a predictive model $F$ is expected to be designed to produce the forecast $Y^{\prime}=F(X)$, aiming to minimize the discrepancy between the predicted $Y^{\prime}$ and the actual $Y$, thereby improving the accuracy of the forecasts.

\subsection{Transformer-based Time Series Forecasting}
Transformer models have become mainstream methods for time series forecasting due to their self-attention mechanism, which captures long-term dependencies without the vanishing or exploding gradient issues of RNNs. However, the canonical Transformer model has quadratic complexity \cite{li2019enhancing}. Informer \cite{zhou2021informer} reduces the complexity to $O(LlogL)$ with a sparse attention mechanism, enhancing performance for long sequences. Autoformer \cite{wu2021autoformer} addresses bottlenecks in sparse attention by proposing a deep decomposition architecture for better information extraction. Crossformer \cite{zhang2023crossformer} introduces dual-phase attention for temporal and variable dimensions, improving multivariate time series handling. PatchTST \cite{nie2023a} segments the input sequence into smaller patches, allowing the model to process longer sequences while reducing computational complexity. iTransformer \cite{liu2024itransformer} modifies the Transformer by inverting the roles of self-attention and feed-forward mechanisms, making it more suitable for time series tasks.

Although complexity has been progressively reduced by various model explorations in Transformer-based time series forecasting, limitations remain due to the inherent computational demands of the self-attention mechanism and the necessity to process complete sequence information. Various improvement efforts have been made, including sparse patterns, kernelization methods, and chunking techniques. However, when managing long sequences, these approaches still encounter the following challenges, including the trade-off between efficiency and accuracy, reliance on extensive training data, and difficulty capturing subtle dependencies within time series. 

%More recently, the emergence of the iTransformer has modified the classic Transformer architecture by inverting the roles of self-attention and the feed-forward neural network (FNN) mechanisms, and by embedding each variable's time series as a token, making it more suited for time series problems.

\subsection{MLP-based Time Series Forecasting}
MLP-based models have high interpretability, low computational costs, and ability to model long inputs \cite{ekambaram2023tsmixer}, while they face challenges to the Transformer series. DLinear \cite{zeng2023transformers} has initially demonstrated that linear models can perform comparably to, if not better, Transformer architectures in time series forecasting, particularly in capturing trends and residuals. It retains its effectiveness with increasing lengths of the retrospective window, contrasting with the poor performance of classic Transformer architecture models. TiDE \cite{das2023longterm} consisting of an encoder and a decoder implemented via multilayer perceptrons, takes advantage of the simplicity and speed of linear models while addressing challenges such as long-term dependencies, noise, and uncertainties inherent in time series forecasting. RLinear \cite{li2023revisiting} proposes a single-layer linear model integrating reversible instance normalization (RevIN) with channel independence, which has demonstrated exceptional predictive capabilities across a variety of datasets. However, according to comparative analyses in the iTransformer literature, the predictive difficulty of linear models increases with longer forecast lengths due to a reduced ratio of input data. 

Although MLP-based models are adept at detecting periodic patterns within extensive input data, real-world scenarios frequently involve prediction windows substantially longer than the inputs. This highlights the need for further research to extend forecasting lengths while maintaining predictive accuracy.

\subsection{SSM-Based Sequence Modeling}
To date, sequence modeling methods based on SSM have garnered increasing attention. SSM represents system dynamics through latent states evolving over time and is renowned for its linear complexity and capacity to handle long sequences. It incorporates both observed data and hidden states, allowing for a comprehensive understanding of temporal dependencies and underlying processes. SSM is adaptable to various forms of sequential data, including those with irregularities and noise. By integrating domain knowledge and handling multivariate sequences, it enhances prediction accuracy. SSM is extensively applied in fields such as economics, engineering, and environmental science, offering robust and interpretable predictions. S4 \cite{gu2021efficiently} (Structured State Space Sequence) combines linear SSMs, the HiPPO framework, and deep learning to achieve high performance. S5 \cite{smith2022simplified} replaces the frequency-domain methods used in S4 with a purely recurrent time-domain approach that leverages parallel scanning. S5 maintains the computational efficiency of S4 while achieving superior performance. Mamba \cite{gu2023mamba}, building on S4, introduces an information-selective mechanism and hardware-aware acceleration algorithms, further enhancing the performance and computational efficiency of sequence modeling. It has achieved success across multiple domains. S-Mamba \cite{wang2024mamba} utilizes a bidirectional Mamba layer to extract inter-variate correlations, while TimeMachine \cite{ahamed2024timemachine} employs an ensemble of four Mamba models, both representing significant advancements in time series researches.

However, studies using Mamba on time series forecasting models have only recently emerged and require further improvements to enhance its ability to extract both short-term and long-term features, as well as to improve prediction accuracy across different channel scenarios. This paper focuses on addressing the extraction of multi-scale temporal features and expanding the predictive capabilities of Mamba across various channel scenarios while maintaining lower memory usage.

\section{The Proposed Method: UmambaTSF}
In this section, we first introduce the overall architecture of the UmambaTSF model. Then, we provide a detailed explanation of the multi-scale feature extractor, with a focus on its core component, the Mamba-based temporal signal processor (MTSP).

\begin{figure*}[h]
\vspace{-1em}
\centering
\includegraphics[width=1\textwidth]{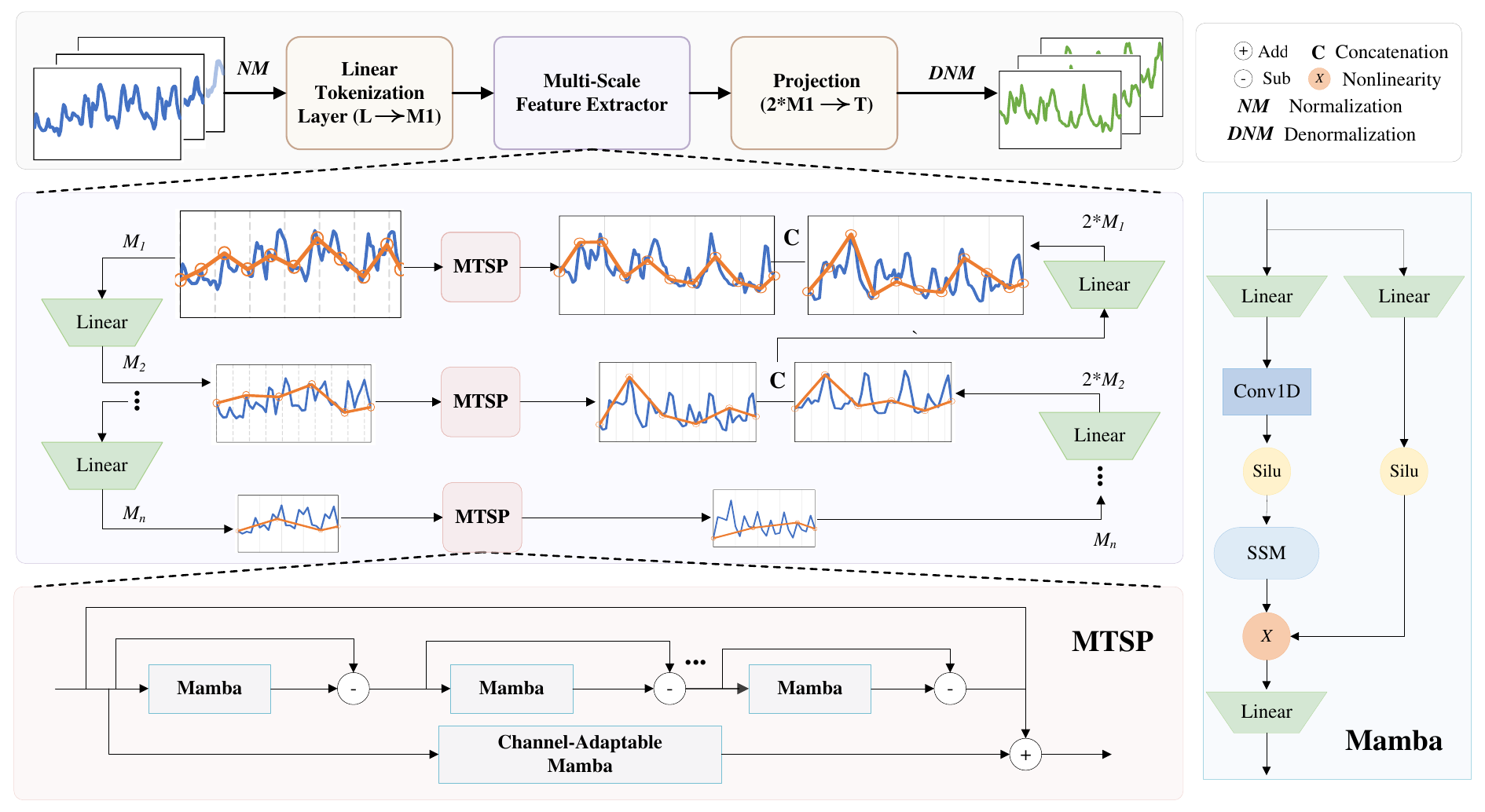}
\caption{Overall framework of UmambaTSF, where the top-left section illustrates the overall architecture of our model. The center-left and bottom-left sections introduce the multi-scale feature extractor and the Mamba-based temporal signal processor (MTSP), which are used to extract temporal features at each scale. The top-right section provides an explanation of the operation symbols, while the bottom-right section describes the structure of the Mamba model.}
\label{fig:overall}
\end{figure*}

\subsection{Architecture Overview}
The architecture of UmambaTSF is illustrated in Fig. \ref{fig:overall}. The time series input data is first processed through instance normalization, utilizing standard Z-score normalization \cite{zhou2023one} and RevIN techniques \cite{kim2022reversible}. This normalization step avoids trend variations due to moment statistics while retaining essential statistical information needed to reconstruct accurate forecasting outcomes. Next, a linear tokenization layer is introduced to map the normalized data to an expanded feature dimension. A multi-scale feature extractor then captures temporal correlations across various scales, after which the multi-scale data is projected to the forecast horizon using a projection layer. Ultimately, instance denormalization is applied to generate the final predictive sequence.

\textbf{The multi-scale feature extractor} is a crucial module within our framework, structured as a U-shaped architecture. This structure facilitates the rich expression of features \cite{ma2024u}. In Fig. \ref{fig:overall}, the left side works as an encoder, utilizing linear layers to map data progressively to shorter lengths of time series, while the right side serves as a decoder, incrementally increasing the dimensional features of temporal data to deeply mine multi-scale characteristics. 

\textbf{The Mamba-based temporal signal processor}, a core component of the multi-scale feature extractor, employs residual Mamba layers and channel-adaptable Mamba block to flexibly and efficiently extract long-range contextual features from each dimensional feature. These components are designed to extract periodic signals and trend information from time series data, while capturing interactions among multivariate elements under different scenario conditions.

\subsection{U-shaped Multi-scale Feature Extractor}
Time series data typically exhibit diverse periodicities and trends. Our predictive model, UmambaTSF, is based on a U-shaped architecture to extract multi-scale features, allowing it to effectively capture complex data fluctuations. 

As illustrated in Fig.~\ref{fig:overall}, prior to the multi-scale feature extraction module, the Linear tokenizer layer linearly maps the input data $X$ with the length of $L$ to a higher feature dimension $M_1$ ($M_1$ is the data dimension of the first layer in the multi-scale feature extractor). In the encoder part of the multi-scale feature extractor, the feature dimensions are progressively reduced through linear layers from $[M_1, M_2, \ldots, M_n]$, where $M_1 > M_2 > \cdots > M_n$, $n$ represents the number of layers in the multi-scale feature extractor.

In the multi-scale feature extractor, $X_i$ is the feature of the $i$-th layer in the downsampling process of the left-side encoder, from top to bottom. $X_m[i]$ is the temporal association at a certain scale, indicating the lateral skip connections of each layer. $X_i$ and $X_m[i]$ can be defined as follows:
\begin{equation}
X_i = DO(Linear(X_{i-1}, M_{i-1} \to M_i)),
\label{eq:enliner}
\end{equation}
\begin{equation}
X_m[i] = \overrightarrow{MTSP}(X_i),
\label{eq:enmtsp}
\end{equation}
where $i \in \{1, 2, 3, \ldots, n\}$. $Linear$ is achieved through MLP. $M_{i-1} \to M_i$ represents the mapping from dimension $M_{i-1}$ to dimension $M_i$. $DO$ stands for the dropout operation. $\overrightarrow{MTSP}$ indicates that the features in each encoder-decoder layer are further processed by the Mamba-based temporal signal processor to extract temporal dependencies at a specific scale. $X_0$ and $M_0$ represent the initial time series $X$ and input length $L$, respectively. 

%Prior to the multi-scale feature extraction module, the Linear tokenizer layer linearly maps the input data $X$ with the length of $L$ to a higher feature dimension $M_1$. In the multi-feature extraction module, the encoder progressively reduces the feature dimensions  through linear layers from $[M_1, M_2, \ldots, M_n]$, according to equations \ref{eq:enliner} and \ref{eq:enmtsp}, where $M_1 > M_2 > \cdots > M_n$ and $n$ represents the number of layers in the multi-scale feature extractor. The features in each encoder-decoder layer are further processed by the Mamba-based temporal signal processor, denoted as $\overrightarrow{MTSP}$, to extract temporal dependencies at a specific scale.

In the decoder part, the feature dimensions progressively increase through a series of expanding linear layers, with dimensions ranging from $[M_n, M_{n-1}, \ldots, M_1]$. $X_j^{\prime}$ represents the feature of the $j$-th layer in the upsampling process of the right-side decoder, from bottom to top. $X_0^{\prime}$ is the value of the last layer $X_m$ in the downsampling on the left side. If $j=1$, $X_j^{\prime}$ is calculated by: 
\begin{equation}
X_1^{\prime}= Linear(X_{0}^{\prime}, M_n \to M_{n-1})
\label{eq:delin1}
\end{equation}

The feature obtained from the decoder layer $X^{\prime}$ is concatenated with the corresponding layer's $X_m$ along the temporal length dimension. Upsampling is then applied to linearly map the concatenated feature to the dimension of the previous layer, as shown in equations (\ref{eq:concat}) and (\ref{eq:delin}): 
\begin{equation}
X_j^{\prime} = Concat(X_j^{\prime}, X_m[n-j]), \\
\label{eq:concat}
\end{equation}
\begin{equation}
X_{j+1}^{\prime} = Linear(X_{j}^{\prime}, 2 * M_{n-j} \to M_{n-j-1}),
\label{eq:delin}
\end{equation}
where $j \in \{1, 2, \ldots, n-1\}$. When $j=n-1$, $M_0$ is the forecast length $T$. The projection layer maps the feature dimension of $2 \times M_1$ obtained from the U-shaped multi-scale feature extractor to the forecast length $T$.

\subsection{Mamba-based Temporal Signal Processor}
The structure of the MTSP is composed of fundamental Mamba blocks, representing a state space modeling method with the capability to capture any cyclic process within latent states. The processor consists of three paths: the first, known as residual Mamba layers, is designed to extract cyclical and trend information from time series data; the second, termed channel-adaptable Mamba, handles the mixed processing of multi-dimensional variables; and the third path directly processes the input time series data. The outputs from the three paths are summed, enabling the acquisition of unique temporal signals at each scale while facilitating the flexible processing of multi-channel data. Next, we introduce the fundamental \textbf{Mamba block}, followed by the \textbf{residual Mamba layers} and the \textbf{channel-adaptable Mamba block}.
\paragraph{\textbf{Mamba Block}}
Within the Mamba block, as shown in the bottom right part of Fig. \ref{fig:overall}, both branches first undergo linear mapping. The first branch then proceeds through the one-dimensional causal convolution and SiLU \cite{elfwing2018sigmoid} activation, followed by the structured SSM. This is combined with the activated residual connection of the second branch. Finally, the output is obtained through a linear transformation. Continuous-time SSM maps an input function or sequence $x(t)$ to output $y(t)$ through the latent state $h(t)$ as presented in equation (\ref{eq:ssm-yt}):
\begin{equation}
\begin{aligned}
dh(t) / dt & =\boldsymbol{A} h(t)+\boldsymbol{B} x(t) \\
y(t) & =\boldsymbol{C} h(t),
\label{eq:ssm-yt}
\end{aligned}    
\end{equation}
where A, B, and C are learnable matrices, and the continuous sequence is discretized using a step size $\Delta$. The discretized SSM model is illustrated in equation (\ref{eq:ssm-yt2}):
\begin{equation}
\begin{aligned}
h_t & =\overline{\boldsymbol{A}} h_{t-1}+\overline{\boldsymbol{B}} x_t \\
y_t & =\boldsymbol{C} h_t
\label{eq:ssm-yt2}
\end{aligned}    
\end{equation}
To perform calculations using continuous recursive methods, discrete form $\overline{\boldsymbol{A}}$ and $\overline{\boldsymbol{B}}$ are obtained based on continuous form $A$ and $B$ by equation (\ref{eq:discrete}):
\begin{equation}
\overline{\boldsymbol{A}}=\exp (\Delta \boldsymbol{A}) \\ \quad \overline{\boldsymbol{B}}=(\Delta \boldsymbol{A})^{-1}(\exp (\Delta \boldsymbol{A})-I) \cdot \Delta \boldsymbol{B}
\label{eq:discrete}
\end{equation}

\begin{figure}
\vspace{-1em}
\centering
\includegraphics[width=0.5\textwidth]{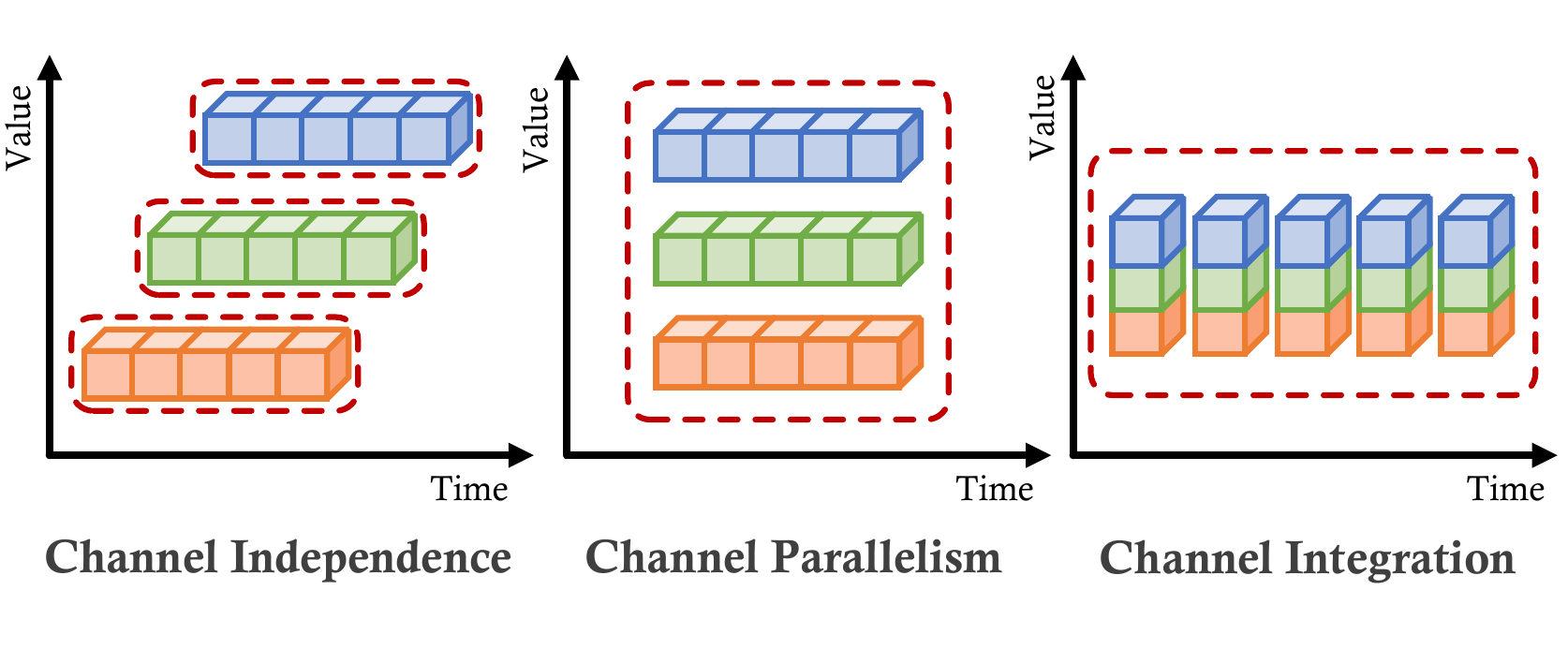}
\vspace{-2em}
\caption{Schematic Diagrams of Channel Processing in Three Different Scenarios.}
\vspace{-0.5em}
\label{fig:channel}
\end{figure}

Additionally, Mamba offers an option to increase the model dimensions using a controllable expansion factor. This enables the model to use coefficient matrices that allow for selective propagation or forgetting of information along the input token sequence, depending on the context and the current token  \cite{gu2023mamba}.
\paragraph{\textbf{Residual Mamba Layers}}
As the residuals can preserve temporal information at different granularities \cite{hou2022multi}, the residuals from multiple Mamba blocks are used to continuously identify the hidden trend information within the temporal features. The input to the first Mamba block is the original input feature, while its output is the reconstructed value corresponding to that input. For the remaining Mamba blocks, the input is the reconstruction value from the previous Mamba block minus the input value of that block. On one hand, the next block complements the previous one by continually fitting the residual information that the earlier block did not capture. On the other hand, this process can be viewed as a decomposition of the time series data, enabling continuous learning of trend information within the temporal dataset. The reconstructed value $rml[k]$ from the $k$-th Mamba block and residual-updated $X_i$ are defined as follows:
\begin{equation}
rml[k] = Mamba(X_i),
\label{eq:rml}
\end{equation}
\begin{equation}
X_i = rml[k] - X_i,
\label{eq:xi}
\end{equation}
where $k \in \{1, 2, \ldots, K\}$, $K$ represents the number of layers in the residual Mamba layer, and each layer is a standard Mamba Block.
\paragraph{\textbf{Channel-Adaptable Mamba Block}}
The channel-adaptable Mamba block consists of a single Mamba designed to process multivariate time series feature information. It is structured as a flexible module to handle various scenarios, including channel independence, channel interdependence, and channel integration. In different scenarios, the input sequences are transformed into various shapes, which correspondingly alter the input dimension $d\_model$ setting within the Mamba block.

As shown in Fig. \ref{fig:channel}, when channels are independent, the dimensions of all time points for each variable are concatenated together and sequentially input as vectors into the Mamba block. When channels are parallel, tokens are formed by variables, with the distinction that multiple variables are input into the Mamba block simultaneously. During channel integration, variables at each time point are concatenated and sequentially fed into the Mamba block. 

For datasets with fewer channels and lower complementarity, the channel independence method enables multivariate predictions unaffected by other variables. Conversely, when datasets have numerous channels with complex interrelations, the channel parallelism method reduces computational costs while incorporating interactions among channels. When multiple channels are strongly interrelated, the channel integration method emphasizes relationships between variables at each time step. 

The input sequence $X_i$ undergoes a transformation based on different channel processing methods, and is represented as $T\_X_i$. The features $cam$ obtained from the channel-adaptable Mamba block must ultimately be subjected to a dimensional transformation to $T\_cam$ to accommodate the summation operation across the three paths. Therefore, $cam$ and $\overrightarrow {MTSP}(X_i)$ are defined as follows:
\begin{equation}
cam = Mamba(T\_X_i)
\end{equation}
\begin{equation}
\overrightarrow {MTSP}(X_i) = rml[K] + T\_cam + X_i
\end{equation}

\section{Experimental Design}
\label{sec:pagestyle}
\subsection{Datasets}
Real-world datasets often contain noise and complex patterns that are absent in synthetic data. Time series modeling on these datasets helps improve the model's capability to handle and predict under realistic and frequently non-ideal conditions. We evaluate our model on seven real-world datasets, including Weather \cite{wu2021autoformer}, Electricity (ECL) \cite{wu2021autoformer}, Traffic \cite{wu2021autoformer}, and four ETT \cite{zhou2021informer} datasets (i.e., ETTh1, ETTh2, ETTm1, ETTm2). These datasets have been widely used for long-term time series forecasting. The Weather dataset comprises 21 meteorological factors, meticulously collected every 10 minutes throughout 2020 from the Weather Station of the Max Planck Biogeochemistry Institute. ECL captures the hourly electricity consumption of 321 clients. The Traffic dataset comprises hourly road occupancy rates recorded by 862 sensors on freeways in the San Francisco Bay Area from January 2015 to December 2016. The ETT dataset includes data on seven factors related to electricity transformers, spanning from July 2016 to July 2018. ETT is divided into four subsets: ETTh1 and ETTh2 (they contain hourly recordings), as well as ETTm1 and ETTm2, which feature recordings every 15 minutes. The relevant statistics for these datasets are shown in TABLE \ref{tab:stat}.

\begin{table}[h]
\caption{The statistical characteristics overview of the seven datasets.}
%\vspace{1em}
\centering
\begin{tabular}{l|l|l|l}
\hline
Dataset & Channels & Time Points & Frequency  \\ \hline
Weather       & 21           & 52696       & 10 Minutes \\
Traffic       & 862          & 17544       & Hourly     \\
Electricity   & 321          & 26304       & Hourly     \\
ETTh1         & 7            & 17420       & Hourly     \\
ETTh2         & 7            & 17420       & Hourly     \\
ETTm1         & 7            & 69680       & 15 Minutes \\
ETTm2         & 7            & 69680       & 15 Minutes \\ \hline
\end{tabular}

\label{tab:stat}
\end{table}

\subsection{Experimental Setting and Metrics}
Similar to most classic time series forecasting models, we divided the dataset into training, validation, and testing segments in a 7:2:1 ratio \cite{wu2021autoformer, liu2024itransformer,dai2024periodicity}. UmambaTSF is implemented using Pytorch and trained on an Ubuntu 20.04.6 system with an Nvidia A100 GPU (40GB). The input sequence length is fixed at $L = 96$, and the forecasting sequence length $T \in \{96, 192, 336, 720\}$. UmambaTSF utilizes the Adam optimizer combined with L2 loss, and the batch size is variable depending on the dataset, but all training sessions are consistent at 20 epochs. UmambaTSF's multi-scale feature extractor layers are set to 2 or 3, with 3 or 4 residual Mamba layers. The state expansion factor of each small Mamba is capped at 16. To ensure model reproducibility, the random seed is fixed. Mean Squared Error (MSE) and Mean Absolute Error (MAE) are employed to assess the effectiveness of long-term predictions. The calculation is shown in equations (\ref{eq:mse}) and (\ref{eq:mae}):
\begin{equation}
MSE  =\frac{\sum_{i=1}^N(Y-\hat{Y})^2}{N}
\label{eq:mse}
\end{equation}

\begin{equation}
MAE  =\frac{1}{N} \sum_{i=1}^N(Y-\hat{Y}),
\label{eq:mae}
\end{equation}
where $N$ represents the number of variables, $Y$ and $Y^{\prime}$ refers to the ground truth and corresponding prediction, respectively.

\subsection{Baseline Methods}
In the experiments, 10 state-of-the-art (SOTA) models are selected as baselines, introduced as follows:

\begin{itemize}
\item[] \textbf{S-Mamba} \cite{wang2024mamba}: It employs a bidirectional Mamba layer to extract inter-variate correlations.
%\newline

\item[] \textbf{iTransformer} \cite{liu2024itransformer}: It adapts the classic Transformer by swapping self-attention and FNN roles and encoding each time series variable as a token.
%\newline

\item[] \textbf{PatchTST} \cite{nie2023a}: It breaks down sequences into smaller patches for efficient processing, using a channel-wise approach.
%\newline

\item[] \textbf{Crossformer} \cite{zhang2023crossformer}: It uses a dual-phase attention mechanism for temporal and variable dimensions, boosting its ability to manage multivariate time series.
%\newline

\item[] \textbf{Autoformer} \cite{wu2021autoformer}: It employs a deep decomposition architecture with $O(L\log L)$ complexity to mitigate sparse attention bottlenecks.
%\newline

\item[] \textbf{FEDformer} \cite{zhou2022fedformer}: It converts time-domain data to frequency-domain, using low-rank approximation to enhance performance and reduce computational complexity.
%\newline

\item[] \textbf{DLinear} \cite{zeng2023transformers}: It shows that linear models can match or surpass Transformer architectures in time series forecasting.
\newline

\item[] \textbf{TiDE} \cite{das2023longterm}: It addresses long-term dependencies, noise, and uncertainties in time series forecasting.
%\newline

\item[] \textbf{RLinear} \cite{li2023revisiting}: It combines a single-layer linear model with RevIN Instance normalization, showing strong predictive performance across multiple datasets.
%\newline

\item[] \textbf{TimesNet} \cite{wu2023timesnet}: It reshapes sequences in two-dimensional space to capture intra-cycle and inter-cycle variations.
%\newline
\end{itemize}	

Among these, the Mamba-based model is S-Mamba; the Transformer-based models include iTransformer, PatchTST, Crossformer, AutoFormer, and FEDformer; the MLP-based models consist of DLinear, TiDE, and RLinear; and the CNN-based method is TimesNet.
 
% Among them, which include the Mamba-based S-Mamba, Transformer-based models such as iTransformer, PatchTST, Crossformer, AutoFormer, and FEDformer, as well as MLP-based methods like DLinear, TiDE, RLinear, and the CNN-based method, i.e., TimesNet. More details about these methods are as follows:

%%%%%%%%%%%%%%table
%\renewcommand{\arraystretch}{2.0}
%\noindent
\setlength{\tabcolsep}{2pt}
\renewcommand{\arraystretch}{1.3}
%\begin{table}[h]
%\centering
%\caption{Full Results in MSE and MAE (the lower the better) for the long-term forecasting task. We compare extensively with baselines under
%different prediction lengths, $L \in {96, 192, 336, 720}$. The length of the input sequence (L) is set to 96 for all baselines. The best results are in highligted in bold red font,and the second best are in underlined bule font.The results of baselines are reported by iTransformer\cite{liu2024itransformer}}
\begin{landscape}
\begin{table}
\caption{Full Results in MSE and MAE (the lower the better) for the long-term forecasting task. We compare extensively with baselines under different prediction lengths, $T \in \{96, 192, 336, 720\}$. The length of the input sequence $L$ is set to 96 for all models. The best results are highlighted in bold red font, and the second best are in underlined blue font. The results of S-Mamba are from S-Mamba \cite{wang2024mamba}, while the results of other baselines are reported by iTransformer \cite{liu2024itransformer}.}
\centering
\begin{tabularx}{\linewidth}
{cl|XX|XX|XX|XX|XX|XX|XX|XX|XX|XX|XX}
\hline
\multicolumn{2}{c|}{Methods}                                                & \multicolumn{2}{c|}{UmambaTS}                         & \multicolumn{2}{c|}{S-Mamba}                                & \multicolumn{2}{c|}{iTransformer}                            & \multicolumn{2}{c|}{Rlinear}                            & \multicolumn{2}{c|}{PatchTST}                       & \multicolumn{2}{c|}{Crossformer}                        & \multicolumn{2}{c|}{TiDE}                          & \multicolumn{2}{c|}{TimesNet}                      & \multicolumn{2}{c|}{Dlinear}                       & \multicolumn{2}{c|}{FEDformer}                     & \multicolumn{2}{c}{Autoformer}                    \\ \hline
D                                              & \multicolumn{1}{c|}{T}     & \multicolumn{1}{c}{MSE}      & \multicolumn{1}{c|}{MAE}     & \multicolumn{1}{c}{MSE}      & \multicolumn{1}{c|}{MAE}     & \multicolumn{1}{c}{MSE}      & \multicolumn{1}{c|}{MAE}      & \multicolumn{1}{c}{MSE} & \multicolumn{1}{c|}{MAE}      & \multicolumn{1}{c}{MSE} & \multicolumn{1}{c|}{MAE} & \multicolumn{1}{c}{MSE}      & \multicolumn{1}{c|}{MAE} & \multicolumn{1}{c}{MSE} & \multicolumn{1}{c|}{MAE} & \multicolumn{1}{c}{MSE} & \multicolumn{1}{c|}{MAE} & \multicolumn{1}{c}{MSE} & \multicolumn{1}{c|}{MAE} & \multicolumn{1}{c}{MSE} & \multicolumn{1}{c|}{MAE} & \multicolumn{1}{c}{MSE} & \multicolumn{1}{c}{MAE} \\ \hline
\multicolumn{1}{c|}{}                          & 96   & \textbf{\textcolor{red}{0.157}} & \textbf{\textcolor{red}{0.204}} & 0.165 & \underline{\textcolor{blue}{0.210}} & 0.174                        & 0.214                        & 0.192                   & 0.232                        & 0.177                   & 0.218                    & \underline{\textcolor{blue}{0.158}} & 0.230                   & 0.202                   & 0.261                   & 0.172                   & 0.220                   & 0.196                   & 0.255                   & 0.217                   & 0.296                   & 0.266                   & 0.336                   \\
\multicolumn{1}{c|}{}                          & 192                        & \textbf{\textcolor{red}{0.205}} & \textbf{\textcolor{red}{0.248}} &  0.214 & \underline{\textcolor{blue}{0.252}} & 0.221                        & 0.254                        & 0.240                   & 0.271                        & 0.225                   & 0.259                    &\textcolor{blue}{0.206} & 0.277                   & 0.242                   & 0.298                   & 0.219                   & 0.261                   & 0.237                   & 0.296                   & 0.276                   & 0.336                   & 0.307                   & 0.367                   \\
\multicolumn{1}{c|}{}                          & 336                        & \textbf{\textcolor{red}{0.251}} & \textbf{\textcolor{red}{0.288}} & 0.274  &  0.297 & 0.278                        & \underline{\textcolor{blue}{0.296}}                       & 0.292                   & 0.307                        & 0.278                   & 0.297                    & \underline{\textcolor{blue}{0.272}} & 0.335                   & 0.287                   & 0.335                   & 0.280                   & 0.306                   & 0.283                   & 0.335                   & 0.339                   & 0.380                   & 0.359                   & 0.395                   \\
\multicolumn{1}{c|}{\multirow{-4}{*}{\rotatebox[origin=c]{90}{Weather}}} & 720                        & \textbf{\textcolor{red}{0.340}} & \textbf{\textcolor{red}{0.344}} &  0.350 & \underline{\textcolor{blue}{0.345}} & 0.358                        & 0.349                        & 0.364                   & 0.353                        & 0.354                   & 0.348                    & 0.398                        & 0.418                   & 0.351                   & 0.386                   & 0.365                   & 0.359                   & \underline{\textcolor{blue}{0.345}}                  & 0.381                   & 0.403                   & 0.428                   & 0.419                   & 0.428                   \\ \hline
\multicolumn{1}{c|}{}                          & 96                         & \underline{\textcolor{blue}{0.391}} & \underline{\textcolor{blue}{0.266}} & \textbf{\textcolor{red}{0.382}} & \textbf{\textcolor{red}{0.261}} & 0.395                        & 0.268 & 0.649                   & 0.389                        & 0.544                   & 0.359                    & 0.522                        & 0.290                   & 0.805                   & 0.493                   & 0.593                   & 0.321                   & 0.650                   & 0.396                   & 0.587                   & 0.366                   & 0.613                   & 0.388                   \\
\multicolumn{1}{c|}{}                          & 192                        & \underline{\textcolor{blue}{0.413}} & \underline{\textcolor{blue}{0.274}} & \textbf{\textcolor{red}{0.396}} & \textbf{\textcolor{red}{0.267}} & 0.417                        & 0.276                        & 0.601                   & 0.366                        & 0.540                   & 0.354                    & 0.530                        & 0.293                   & 0.756                   & 0.474                   & 0.617                   & 0.336                   & 0.598                   & 0.370                   & 0.604                   & 0.373                   & 0.616                   & 0.382                   \\
\multicolumn{1}{c|}{}                          & 336                        & \underline{\textcolor{blue}{0.431}} & \underline{\textcolor{blue}{0.282}} & \textbf{\textcolor{red}{0.417}} & \textbf{\textcolor{red}{0.276}} & 0.433                        & 0.283                        & 0.609                   & 0.369                        & 0.551                   & 0.358                    & 0.558                        & 0.305                   & 0.762                   & 0.477                   & 0.629                   & 0.336                   & 0.605                   & 0.373                   & 0.621                   & 0.383                   & 0.622                   & 0.337                   \\
\multicolumn{1}{c|}{\multirow{-4}{*}{\rotatebox[origin=c]{90}{Traffic}}} & 720                        & \underline{\textcolor{blue}{0.466}} & \underline{\textcolor{blue}{0.302}} & \textbf{\textcolor{red}{0.460}} & \textbf{\textcolor{red}{0.300}} & 0.467                        & \underline{\textcolor{blue}{0.302}} & 0.647                   & 0.387                        & 0.586                   & 0.375                    & 0.589                        & 0.328                   & 0.719                   & 0.449                   & 0.640                   & 0.350                   & 0.645                   & 0.394                   & 0.626                   & 0.382                   & 0.660                   & 0.408                   \\ \hline
\multicolumn{1}{c|}{}                          & 96                         & \underline{\textcolor{blue}{0.140}} & \underline{\textcolor{blue}{0.236}} & \textbf{\textcolor{red}{0.139}} & \textbf{\textcolor{red}{0.235}} & 0.148                        & 0.240                        & 0.201                   & 0.281                        & 0.195                   & 0.285                    & 0.219                        & 0.314                   & 0.237                   & 0.329                   & 0.168                   & 0.272                   & 0.197                   & 0.282                   & 0.193                   & 0.308                   & 0.201                   & 0.317                   \\
\multicolumn{1}{c|}{}                          & 192                        & \textbf{\textcolor{red}{0.157}} & \textbf{\textcolor{red}{0.251}} & \underline{\textcolor{blue}{0.159}} & 0.255 & 0.162                        & \underline{\textcolor{blue}{0.253}} & 0.201                   & 0.283                        & 0.199                   & 0.289                    & 0.231                        & 0.322                   & 0.236                   & 0.330                   & 0.184                   & 0.289                   & 0.196                   & 0.285                   & 0.201                   & 0.315                   & 0.222                   & 0.334                   \\
\multicolumn{1}{c|}{}                          & 336                        & \textbf{\textcolor{red}{0.173}} & \textbf{\textcolor{red}{0.268}} & \underline{\textcolor{blue}{0.176}} & 0.272 & 0.178                        & \underline{\textcolor{blue}{0.269}}                       & 0.215                   & 0.298                        & 0.215                   & 0.305                    & 0.246                        & 0.337                   & 0.249                   & 0.344                   & 0.198                   & 0.300                   & 0.209                   & 0.301                   & 0.214                   & 0.329                   & 0.231                   & 0.338                   \\
\multicolumn{1}{c|}{\multirow{-4}{*}{\rotatebox[origin=c]{90}{ECL}}}     & 720 & \textbf{\textcolor{red}{0.203}} & \textbf{\textcolor{red}{0.296}} & \underline{\textcolor{blue}{0.204}} & \underline{\textcolor{blue}{0.298}} & 0.225                        & 0.317                        & 0.257                   & 0.331                        & 0.256                   & 0.337                    & 0.280                        & 0.363                   & 0.284                   & 0.373                   & 0.220                   & 0.320                   & 0.245                   & 0.333                   & 0.246                   & 0.355                   & 0.254                   & 0.361                   \\ \hline
\multicolumn{1}{c|}{}                          &  96  & \textbf{\textcolor{red}{0.358}} & \textbf{\textcolor{red}{0.391}} & 0.386                        & 0.405                        & 0.386                        & 0.405                        & 0.386                   & \underline{\textcolor{blue}{0.395}}                       & 0.414                   & 0.419                    & 0.423                        & 0.448                   & 0.479                   & 0.464                   & 0.384                   & 0.402                   & 0.386                   & 0.400                   & \underline{\textcolor{blue}{0.376}}                   & 0.419                   & 0.449                   & 0.459                   \\
\multicolumn{1}{c|}{}                          & 192 & \textbf{\textcolor{red}{0.415}} & \textbf{\textcolor{red}{0.420}} & 0.443                        & 0.437                        & 0.441                        & 0.436                        & 0.437                   & \underline{\textcolor{blue}{0.424}} & 0.460                   & 0.445                    & 0.471                        & 0.474                   & 0.525                   & 0.492                   & 0.436                   & 0.429                   & 0.437                   & 0.432                   &  \underline{\textcolor{blue}{0.420}}                   & 0.448                   & 0.500                   & 0.482                   \\
\multicolumn{1}{c|}{}                          & 336 & \textbf{\textcolor{red}{0.428}} & \textbf{\textcolor{red}{0.426}} & 0.489                        & 0.468                        & 0.487                        & 0.458                        & 0.479                   & \underline{\textcolor{blue}{0.446}}                        & 0.501                   & 0.466                    & 0.570                        & 0.546                   & 0.565                   & 0.515                   & 0.491                   & 0.469                   & 0.481                   & 0.459                 & \underline{\textcolor{blue}{0.459}}                   & 0.465                   & 0.521                   & 0.496                   \\
\multicolumn{1}{c|}{\multirow{-4}{*}{\rotatebox[origin=c]{90}{ETTh1}}}   & 720 & \textbf{\textcolor{red}{0.454}} & \textbf{\textcolor{red}{0.456}} & 0.502                        & 0.489                        & 0.503                        & 0.491                        & \underline{\textcolor{blue}{0.481}}                   & \underline{\textcolor{blue}{0.470}}                        & 0.500                   & 0.488                    & 0.653                        & 0.621                   & 0.594                   & 0.558                   & 0.521                   & 0.500                   & 0.519                   & 0.516                   & 0.506                   & 0.507                   & 0.514                   & 0.512                   \\ \hline
\multicolumn{1}{c|}{}                          & 96 & \textbf{\textcolor{red}{0.274}} & \textbf{\textcolor{red}{0.334}} & 0.296                        & 0.348                        & 0.297                        & 0.349                        & \underline{\textcolor{blue}{0.288}}                   & \underline{\textcolor{blue}{0.338}}                        & 0.302                   & 0.348                    & 0.745                        & 0.584                   & 0.400                   & 0.440                   & 0.340                   & 0.374                   & 0.333                   & 0.387                   & 0.358                   & 0.397                   & 0.346                   & 0.388                   \\
\multicolumn{1}{c|}{}                          &192 & \textbf{\textcolor{red}{0.347}} & \textbf{\textcolor{red}{0.381}} & 0.376                        & 0.396                        & 0.381 & 0.400                        & \underline{\textcolor{blue}{0.374}}                   & \underline{\textcolor{blue}{0.390}}                        & 0.388                   & 0.400                    & 0.877                        & 0.656                   & 0.528                   & 0.509                   & 0.402                   & 0.414                   & 0.477                   & 0.476                   & 0.429                   & 0.439                   & 0.456                   & 0.452                   \\
\multicolumn{1}{c|}{}                          & 336 & \textbf{\textcolor{red}{0.340}} & \textbf{\textcolor{red}{0.381}} & 0.424                        & 0.431                        & 0.428                        & 0.432                        & \underline{\textcolor{blue}{0.415}}                   & \underline{\textcolor{blue}{0.426}}                        & 0.426                   & 0.433                    & 1.043                        & 0.731                   & 0.643                   & 0.571                   & 0.452                   & 0.452                   & 0.594                   & 0.541                   & 0.496                   & 0.487                   & 0.482                   & 0.486                   \\
\multicolumn{1}{c|}{\multirow{-4}{*}{\rotatebox[origin=c]{90}{ETTh2}}}   & 720 & \textbf{\textcolor{red}{0.409}} & \textbf{\textcolor{red}{0.432}} & 0.426                        & 0.444                        & 0.427                        & 0.445                        & \underline{\textcolor{blue}{0.420}}                   & \underline{\textcolor{blue}{0.440}}                        & 0.431                   & 0.446                    & 1.104                        & 0.763                   & 0.874                   & 0.679                   & 0.462                   & 0.468                   & 0.831                   & 0.657                   & 0.463                   & 0.474                   & 0.515                   & 0.511                   \\ \hline
\multicolumn{1}{c|}{}                          & 96                         & \textbf{\textcolor{red}{0.316}} & \textbf{\textcolor{red}{0.356}} & 0.333                        & 0.368                        & 0.334                        & 0.368                        & 0.355                   & 0.376                        & \underline{\textcolor{blue}{0.329}}                   & \underline{\textcolor{blue}{0.367}}                    & 0.404                        & 0.426                   & 0.364                   & 0.387                   & 0.338                   & 0.375                   & 0.345                   & 0.372                   & 0.379                   & 0.419                   & 0.505                   & 0.475                   \\
\multicolumn{1}{c|}{}                          & 192                        & \textbf{\textcolor{red}{0.356}} & \textbf{\textcolor{red}{0.378}} & 0.376                        & 0.390                        &  0.377 & 0.391                        & 0.391                   & 0.392                        & \underline{\textcolor{blue}{0.367}}                   & \underline{\textcolor{blue}{0.385}}                    & 0.450                        & 0.451                   & 0.398                   & 0.404                   & 0.374                   & 0.387                   & 0.380                   & 0.389                   & 0.426                   & 0.441                   & 0.553                   & 0.496                   \\
\multicolumn{1}{c|}{}                          & 336                        & \textbf{\textcolor{red}{0.374}} & \textbf{\textcolor{red}{0.400}} & 0.408                        & 0.413                        & 0.426                        & 0.420                        & 0.424                   & 0.415                        & \underline{\textcolor{blue}{0.399}}             & \underline{\textcolor{blue}{0.410}}                    & 0.532                        & 0.515                   & 0.428                   & 0.425                   & 0.410                   & 0.411                   & 0.413                   & 0.413                   & 0.445                   & 0.459                   & 0.621                   & 0.537                   \\
\multicolumn{1}{c|}{\multirow{-4}{*}{\rotatebox[origin=c]{90}{ETTm1}}}   & 720                        & \textbf{\textcolor{red}{0.440}} & \textbf{\textcolor{red}{0.435}} & 0.475                        & 0.448                        & 0.491                        & 0.459                        & 0.487                   & 0.450                        & \underline{\textcolor{blue}{0.454}}                  & \underline{\textcolor{blue}{0.439}}                    & 0.666                        & 0.589                   & 0.487                   & 0.461                   & 0.478                   & 0.450                   & 0.474                   & 0.453                   & 0.543                   & 0.490                   & 0.671                   & 0.561                   \\ \hline
\multicolumn{1}{c|}{}                          & 96                         & \textbf{\textcolor{red}{0.175}} & \textbf{\textcolor{red}{0.256}} & \underline{\textcolor{blue}{0.179}}                        & 0.263                        & 0.180                        & 0.264                        & 0.182                   & 0.265                        & \textbf{\textcolor{red}{0.175}}                   & \underline{\textcolor{blue}{0.259}}                    & 0.287                        & 0.366                   & 0.207                   & 0.305                   & 0.187                   & 0.267                   & 0.193                   & 0.292                   & 0.203                   & 0.287                   & 0.255                   & 0.339                   \\
\multicolumn{1}{c|}{}                          & 192                        & \textbf{\textcolor{red}{0.238}} & \textbf{\textcolor{red}{0.298}} & 0.250                        & 0.309                        & 0.250                        & 0.309                        & 0.246                   & 0.304                        & \underline{\textcolor{blue}{0.241}}                   & \underline{\textcolor{blue}{0.302}}                    & 0.414                        & 0.492                   & 0.290                   & 0.364                   & 0.249                   & 0.309                   & 0.284                   & 0.362                   & 0.269                   & 0.328                   & 0.281                   & 0.340                   \\
\multicolumn{1}{c|}{}                          & 336                        & \textbf{\textcolor{red}{0.286}} & \textbf{\textcolor{red}{0.331}} & 0.312                        & 0.349                        & 0.311                        & 0.348                        & 0.307                   & \underline{\textcolor{blue}{0.342}}                        & \underline{\textcolor{blue}{0.305}}                   & 0.343                    & 0.597                        & 0.542                   & 0.377                   & 0.422                   & 0.321                   & 0.351                   & 0.369                   & 0.427                   & 0.325                   & 0.366                   & 0.339                   & 0.372                   \\
\multicolumn{1}{c|}{\multirow{-4}{*}{\rotatebox[origin=c]{90}{ETTm2}}}   & 720                        & \textbf{\textcolor{red}{0.370}} & \textbf{\textcolor{red}{0.383}} & 0.411                        & 0.406                        & 0.412                        & 0.407                        & 0.407                   & \underline{\textcolor{blue}{0.398}}                        & \underline{\textcolor{blue}{0.402}}                   & 0.400                    & 1.730                        & 1.042                   & 0.558                   & 0.524                   & 0.408                   & 0.403                   & 0.554                   & 0.522                   & 0.421                   & 0.415                   & 0.433                   & 0.432                   \\ \hline
\end{tabularx}
\label{tab:allres}
\end{table}
\end{landscape}

\section{Results and Analysis}
\subsection{Main Results}
TABLE \ref{tab:allres} presents the overall forecasting performance of UmambaTSF and the baseline models across the seven datasets. Based on the results in the table, we summarize the following observations and provide a detailed analysis:

(1). UmambaTSF outperforms existing SOTA models, including iTransformer and S-Mamba, on the Weather, ECL, ETTh1, ETTh2, ETTm1, and ETTm2 datasets. As shown in TABLE \ref{tab:improv}, UmambaTSF achieves improvements in MSE of 7.75\% and 5.18\% over iTransformer and S-Mamba, respectively on the Weather dataset, along with improvements of 5.62\% and 6.15\% on the ECL dataset. Furthermore, on the four ETT datasets, the average improvements are 8.79\% and 8.20\%, respectively.

(2). UmambaTSF significantly exceeds existing SOTA models on both the channel-rich Weather and ECL datasets as well as the channel-sparse ETT datasets, demonstrating improvements in MSE and MAE. This enhancement is not only attributed to the multi-scale feature extractor's ability to capture temporal dependencies from multiple periodic signals but also to the flexible channel processing techniques provided by the proposed channel-adaptable approach in UmambaTSF. This accommodates transformations across various data scenarios and enhances the model's generalizability.

(3). UmambaTSF surpasses the results of iTransformer on the Traffic dataset but falls short of S-Mamba's performance. This is because UmambaTSF is designed to be highly minimalist and computationally efficient, accommodating both low and high channel counts. For the Traffic dataset, where directly using parallel channel methods involves too much noisy information, we plan to further design channel data filtering capabilities in future work.

\begin{figure*}[h]
    \centering
    \captionsetup[subfloat]{font=footnotesize}
    \subfloat[Forecasting comparisons between UmambaTSF and S-Mamba]{
        \includegraphics[width=0.5\textwidth]{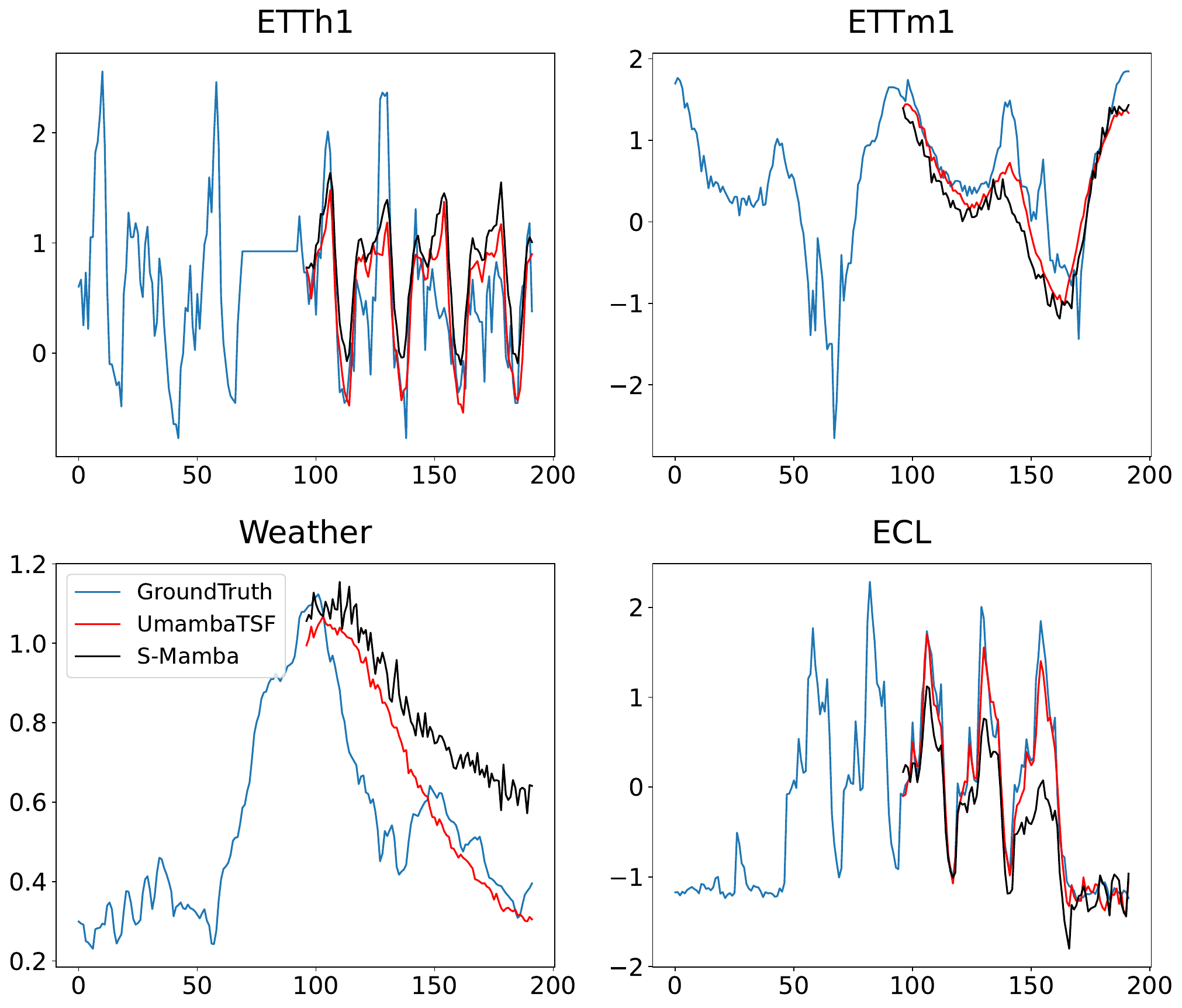}
        }
    %\hfill
    \subfloat[Forecasting comparisons between UmambaTSF and iTransformer]{
        \includegraphics[width=0.5\textwidth]{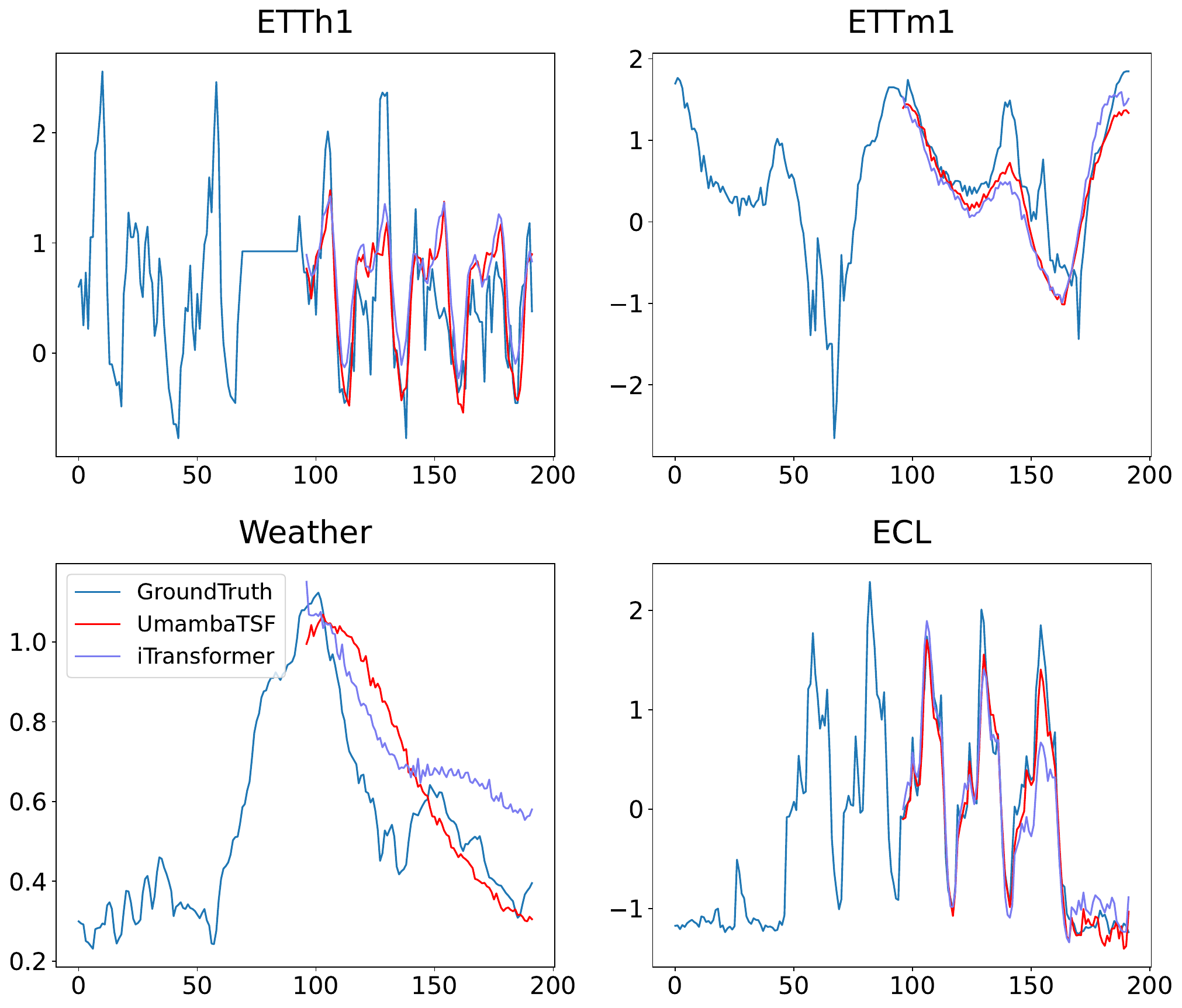}
    }
    \caption{Comparison of forecasts among UmambaTSF, iTransformer, and S-Mamba on four datasets, using an input length of 96 and a forecast length of 96. The blue line depicts the ground truth, while the red, black, and purple lines represent the predictions from UmambaTSF, S-Mamba, and iTransformer, respectively.}
    \label{fig:showcase}
\end{figure*}

\begin{table*}
\caption{The improvement of UmambaTSF over iTransformer and S-Mamba, evaluated using the average MSE and MAE for forecast lengths $T \in \{96, 192, 336, 720\}$ with an input length of $L = 96$.}
\centering
\begin{tabular}{>{\centering\arraybackslash}p{2.2cm}|>{\centering\arraybackslash}p{1.2cm}>{\centering\arraybackslash}p{1.2cm}|>{\centering\arraybackslash}p{1.2cm}>{\centering\arraybackslash}p{1.2cm}|>{\centering\arraybackslash}p{1.2cm}>{\centering\arraybackslash}p{1.2cm}|>{\centering\arraybackslash}p{1.6cm}>{\centering\arraybackslash}p{1.6cm}|>{\centering\arraybackslash}p{1.6cm}>{\centering\arraybackslash}p{1.6cm}}
\noalign{\global\setlength{\arrayrulewidth}{1.0pt}} 
\hline
\noalign{\global\setlength{\arrayrulewidth}{0.1pt}}
\rule{0pt}{4ex} 
Models & \multicolumn{2}{c|}{\makecell{iTransformer \\ \cite{liu2024itransformer}}} & \multicolumn{2}{c|}{\makecell{S-Mamba \\ \cite{wang2024mamba}}} & \multicolumn{2}{c|}{\makecell{UmambaTSF \\ \textbf{(ours)}}} & \multicolumn{2}{c|}{\makecell{Promotion over \\ iTransformer}} & \multicolumn{2}{c}{\makecell{Promotion over\\ S-Mamba}} \\ 
\cmidrule(lr){2-3} \cmidrule(lr){4-5} \cmidrule(lr){6-7} \cmidrule(lr){8-9} \cmidrule(lr){10-11} 
Metric & MSE             & MAE             & MSE           & MAE          & MSE            & MAE           & MSE                & MAE               & MSE             & MAE            \\ 
\noalign{\global\setlength{\arrayrulewidth}{1.0pt}} 
\hline
\noalign{\global\setlength{\arrayrulewidth}{0.1pt}} 
Weather & 0.258           & 0.278           & 0.251         & 0.276        & 0.238          & 0.271         & 7.75\% \(\uparrow\)            & 2.52\% \(\uparrow\)            & 5.18\% \(\uparrow\)         & 1.81\% \(\uparrow\)        \\ \hline
Traffic & 0.428           & 0.282           & 0.414         & 0.276        & 0.425          & 0.281         & 0.70\% \(\uparrow\)            & 0.35\% \(\uparrow\)           & -2.66\%  \(\downarrow\)       & -1.81\% \(\downarrow\)       \\ \hline
ECL     & 0.178           & 0.270           & 0.179         & 0.265        & 0.168          & 0.263         & 5.62\% \(\uparrow\)            & 2.59\%  \(\uparrow\)          & 6.15\% \(\uparrow\)          & 0.75\% \(\uparrow\)        \\ \hline
ETTh1   & 0.454           & 0.448           & 0.455         & 0.450        & 0.414          & 0.423         & 8.81\%  \(\uparrow\)           & 5.58\%  \(\uparrow\)          & 9.01\% \(\uparrow\)          & 6.00\% \(\uparrow\)        \\ \hline
ETTh2   & 0.383           & 0.407           & 0.381         & 0.405        & 0.343          & 0.382         & 10.44\% \(\uparrow\)           & 6.14\% \(\uparrow\)           & 9.97\% \(\uparrow\)         & 5.68\% \(\uparrow\)        \\ \hline
ETTm1   & 0.407           & 0.410           & 0.398         & 0.405        & 0.372          & 0.392         & 8.60\% \(\uparrow\)           & 4.39\%  \(\uparrow\)          & 6.53\% \(\uparrow\)         & 3.21\% \(\uparrow\)        \\ \hline
ETTm2   & 0.288           & 0.332           & 0.288         & 0.332        & 0.267          & 0.317         & 7.29\% \(\uparrow\)            & 4.52\% \(\uparrow\)           & 7.29\% \(\uparrow\)         & 4.52\% \(\uparrow\)        \\ 
\noalign{\global\setlength{\arrayrulewidth}{1.0pt}} 
\hline
\noalign{\global\setlength{\arrayrulewidth}{0.1pt}} 
\end{tabular}

\label{tab:improv}
\end{table*}
Additionally, to more intuitively evaluate the predictive capability of UmambaTSF, we visually compare UmambaTSF with leading baseline models S-Mamba and iTransformer across four datasets (ETTh1, ETTm1, Weather, and ECL) through graphical representation. We randomly select a variable and input its lookback sequence to showcase the predicted results alongside the actual sequence trend. As shown in Fig. \ref{fig:showcase}, the actual input and subsequent sequences are represented by a blue line, while the predictions from UmambaTSF, S-Mamba, and iTransformer are represented by red, black, and purple lines, respectively. It is evident that UmambaTSF's predictions closely align with the actual values, showing nearly perfect consistency on the ECL dataset, a significant advantage on the Weather dataset, and improved alignment with the extremes within the periodic sequences of the ETT datasets.

In time series prediction, time complexity is a crucial factor because it directly affects the scalability and efficiency of the model. Time complexity determines the computational time required to process data, which is particularly vital for real-time or near-real-time applications. The proposed method, UmambaTSF, leverages both the linearly scalable linear layers and the Mamba structure, achieving a time complexity of $O(L)$. This efficiency is crucial for applications that require immediate responses, such as stock trading and weather forecasting, where the model must process data rapidly. However, the mainstream algorithms for time series are still based on the transformer structure. The original Transformer model \cite{vaswani2017attention}, relying on full self-attention mechanisms, has a time complexity of $O(L^2)$, significantly increasing computational costs for long sequence data. To address this challenge, Informer \cite{zhou2021informer} employs sparse self-attention techniques, reducing necessary computations through probabilistic sparsification and entropy coding, thus lowering the time complexity to $O(LlogL)$. PatchTST \cite{nie2023a} segments time series data into blocks and applies the Transformer structure to each block, effectively distributing the computational load and enhancing processing efficiency, despite a time complexity of $O((L/p)^2)$, where $p$ denotes the patch size. These innovations are particularly crucial for large-scale and real-time prediction needs, each balancing computational resource consumption and predictive performance in different ways. 

\begin{figure}[h]
    \centering
    \captionsetup[subfloat]{font=footnotesize}
    \subfloat[ETTh1]{
        \includegraphics[width=0.45\textwidth]{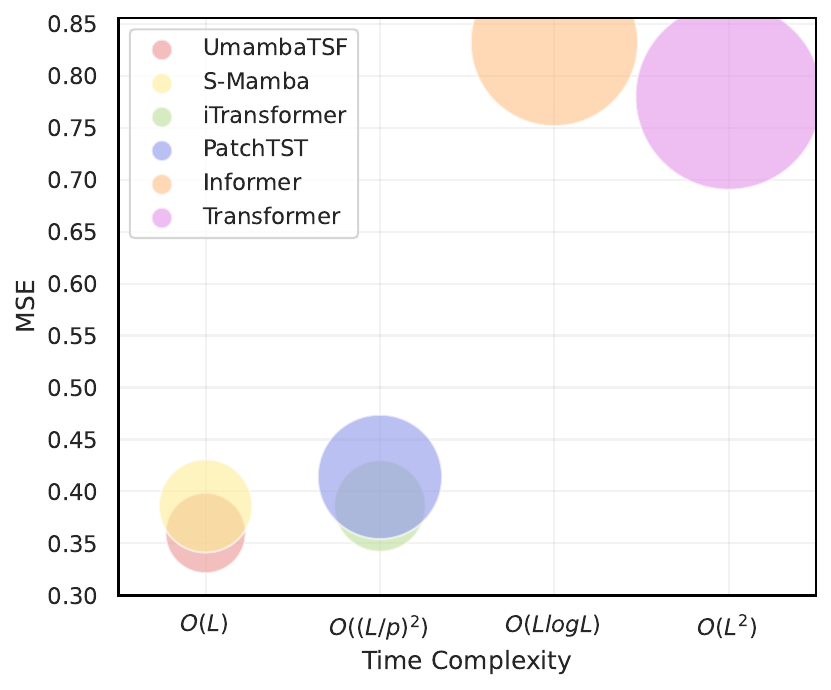}
    }
    \hspace{1mm}
    \subfloat[Weather]{
        \includegraphics[width=0.45\textwidth]{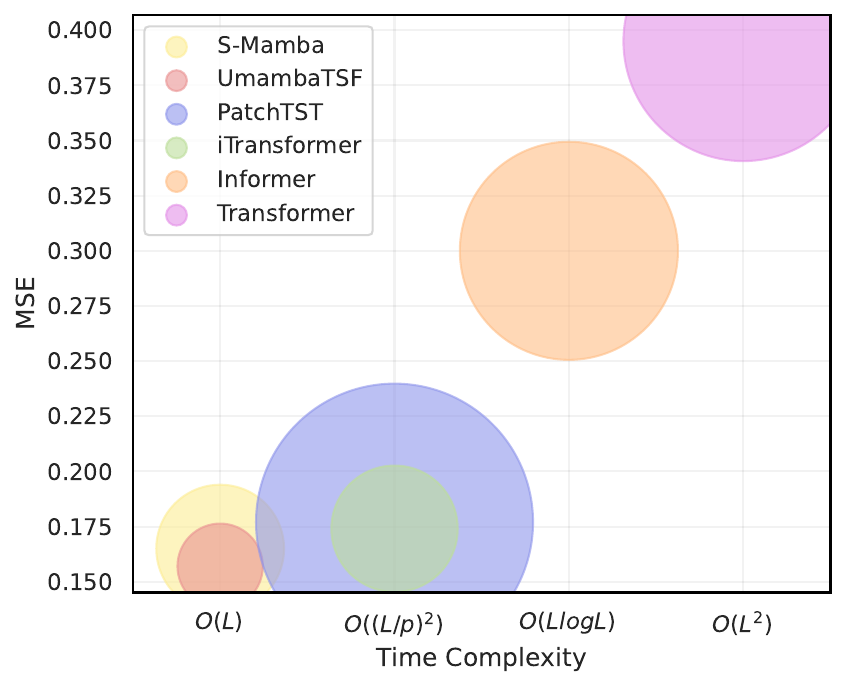}
    }
    \caption{Comparison of UmambaTSF and five models on MSE, Computational Complexity, and GPU Memory. The diameters of the circles represent the memory sizes.}
    \label{fig:advantage}
\end{figure}

Fig. \ref{fig:advantage} presents a comparison of the prediction accuracy, time complexity, and GPU memory usage of UmambaTSF, S-Mamba, iTransformer, PatchTST, Informer, and Transformer models on the ETTh1 and Weather datasets. The specific memory usage values are shown in TABLE \ref{tab:mem}. Note that the complexities $O((L/p)^2)$ and 
$O(LlogL)$ are presented here just for an illustrative purposes. As shown in the figure, UmambaTSF achieves commendable results in both accuracy and computational cost with its purely linear complexity.

\begin{table}[]
\caption{Memory usage (Gib) of UmambaTSF, S-Mamba, and iTransformer when both input length and forecast length are 96.}
\centering
\begin{tabular}{>{\centering\arraybackslash}m{2cm} >{\centering\arraybackslash}m{2cm} >{\centering\arraybackslash}m{2cm} >{\centering\arraybackslash}m{2cm}}
\hline
Model   & UmambaTSF & S-Mamba & iTransformer \\ \hline
ETTh1   & 1.48      & 1.98    & 1.93         \\
Weather & 1.67      & 3.77    & 3.62         \\
ECL     & 1.78      & 2.51    & 2.61         \\ \hline
\end{tabular}

\label{tab:mem}
\end{table}

\subsection{Ablation Study}
To better evaluate the functionality and effectiveness of each module in UmambaTSF, we divide the Multi-Scale Feature Extractor into three components: 1) U-shape Linear Layers (ULL); 2) Residual Mamba Layers (RML); and 3) Channel Adaptable Mamba Block (CAM). As shown in TABLE \ref{tab:ablation}, the “×" and “\checkmark" indicate the absence or presence of a specific module, respectively. When ULL is marked with an “*”, it indicates that both the encoding and decoding layers of ULL consist of only one layer. When RML is marked with an “*”, it signifies that the RML contains only one Mamba block. Due to the coupling relationships between the model components, the RML and CAM modules cannot exist independently of ULL, unless all three modules are absent. All ``×"s represent the model without the Multi-Scale Feature Extractor, while all ``\checkmark"s indicate the complete UmambaTSF model.

\begin{table}[h]
\caption{Results of the ablation study on ETTh1, ETTm1 and Weather datasets. MSE and MAE are the average values across forecast lengths $T$.}
\centering
\begin{tabular}{m{0.6cm}m{0.6cm}m{0.6cm}|m{0.8cm}m{0.8cm}|m{0.8cm}m{0.8cm}|m{0.8cm}m{0.8cm}}
\noalign{\global\setlength{\arrayrulewidth}{1.2pt}} 
\hline
\noalign{\global\setlength{\arrayrulewidth}{0.1pt}} 
\multicolumn{1}{c|}{\multirow{2}{*}{\textbf{ULL}}} & \multicolumn{1}{c|}{\multirow{2}{*}{\textbf{RML}}} & \multicolumn{1}{c|}{\multirow{2}{*}{\textbf{CAM}}} & \multicolumn{2}{c|}{Weather}                          & \multicolumn{2}{c|}{ETTh1}                            & \multicolumn{2}{c}{ETTm1}                             \\ \cmidrule(lr){4-5} \cmidrule(lr){6-7} \cmidrule(lr){8-9}  
\multicolumn{1}{c|}{}                             & \multicolumn{1}{c|}{}                              &                               & MSE                       & \multicolumn{1}{l|}{MAE}  & MSE                       & \multicolumn{1}{l|}{MAE}  & MSE                       & MAE                       \\ 
\noalign{\global\setlength{\arrayrulewidth}{1.2pt}}
\hline
\noalign{\global\setlength{\arrayrulewidth}{0.1pt}}
×                                                 & ×                                                  & ×                             & 0.266                     & 0.290                      & 0.432                     & 0.430                      & 0.410                      & 0.403                     \\
\checkmark                                                 & \checkmark                                                 & ×                             & 0.248 & 0.277 & 0.414 & 0.423 & 0.373 & 0.393 \\
\checkmark                                                 & ×                                                  & \checkmark                             & 0.244                     & 0.274                     & 0.430                     & 0.428                     & 0.408                     & 0.403                     \\
\checkmark                                                 & ×                                                  & ×                             & 0.266                     & 0.29                      & 0.532                     & 0.490                      & 0.520                      & 0.467                     \\
*                                                & \checkmark                                                  & \checkmark                             & 0.241 & 0.274 & 0.423 & 0.427 & 0.372 & 0.392 \\ 
\checkmark                                                 & *                                                 & \checkmark                             & 0.245 & 0.276 & 0.428 & 0.430 & 0.379 & 0.397 \\ 

\checkmark                                                 & \checkmark                                                  & \checkmark                             & 0.238 & 0.271 & 0.414 & 0.423 & 0.372 & 0.392 \\ 
\noalign{\global\setlength{\arrayrulewidth}{1.2pt}} 
\hline
\noalign{\global\setlength{\arrayrulewidth}{0.1pt}} 
\end{tabular}

\label{tab:ablation}
\end{table}

Across multiple experiments, UmambaTSF consistently delivered the best results, with each module contributing uniquely to its overall performance. Notably, the CAM module significantly improved results on the weather dataset, where a high degree of inter-variable correlation is present. On the other hand, the RML module proved particularly important for the ETTh1 and ETTm1 datasets, due mainly to the relatively small number of channels in the ETT series, where the temporal feature information of each channel plays a more critical role.

The model that includes only the ULL module, without RML and CAM, and lacks the concatenation operation in the decoder on the right side of ULL, performs the worst. This indicates that multi-scale representation alone, without detailed temporal signals at each scale, is detrimental to the predictive model. However, when the U-shape structure contains only one layer that includes RML and CAM, or when the RML consists of just one Mamba block, the model still produces reasonable results, though these are not as strong as those achieved by the full UmambaTSF. This demonstrates that the multi-layer feature extraction and the method of using multiple Mamba blocks for residual learning effectively enhance the temporal information, improving performance.

\subsection{Hyperparameter Sensitivity Analysis}
\begin{figure}[h]
    \centering
    \captionsetup[subfloat]{font=footnotesize}
    \subfloat[ETTh1]{
        \includegraphics[width=0.45\textwidth]{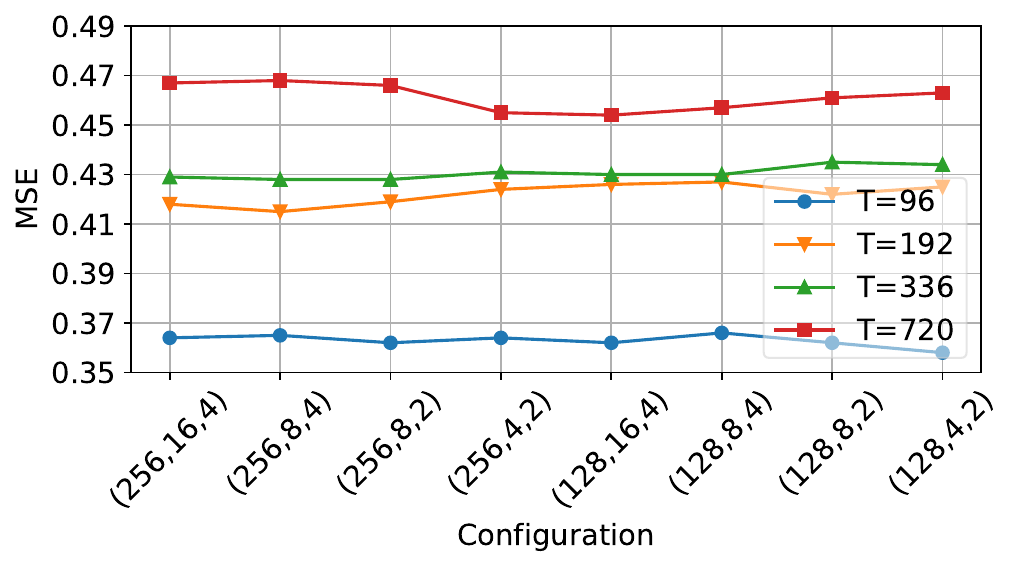}
    }
    \hspace{1mm}
    \subfloat[ETTh2]{
        \includegraphics[width=0.45\textwidth]{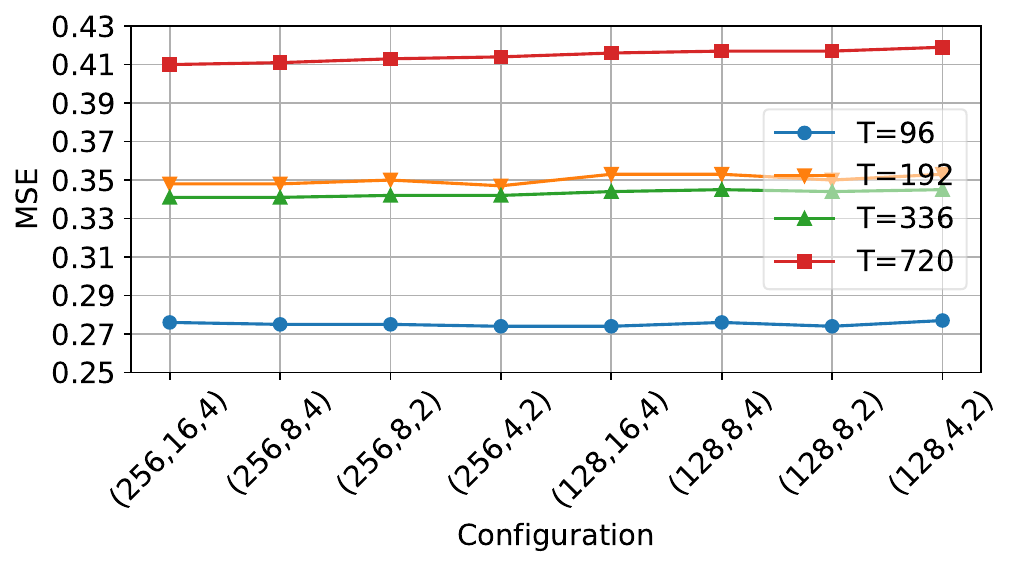}
    }
    \caption{MSE comparison with combinations of multi-scale feature dimensions for input 
sequence length $L = 96$ and forecast length $T \in \{96, 192, 336, 720\}$ for the ETTh1 and ETTh2 datasets. }
 \label{fig:configs}
\end{figure}

As illustrated in the overall framework depicted in Fig. \ref{fig:overall}, our multi-scale feature extraction architecture is primarily composed of linear layers. The encoder begins with a dimension being larger than the input sequence length $L$ and progressively reduces the length. The decoder's scale incrementally rises and merges the temporal dependencies extracted from the same layer by the mamba block. To assess the impact of multi-scale configurations, we conduct experiments on the ETTh1 and ETTh2 datasets with various scale settings, as shown in Fig. \ref{fig:configs}. The results reveal that model performance remains relatively stable across different configurations, with both datasets achieving strong performance under multiple settings. This suggests that once the U-shaped architecture is established, there are several viable scale combinations, and prediction accuracy is not reliant on any single specific configuration.

\subsection{The Generality of UmambaTSF}
\subsubsection{Increasing Lookback Length}
\begin{figure}[h]
    \centering
    \captionsetup[subfloat]{font=footnotesize}
    \subfloat[ETTh1]{
        \includegraphics[width=0.4\textwidth]{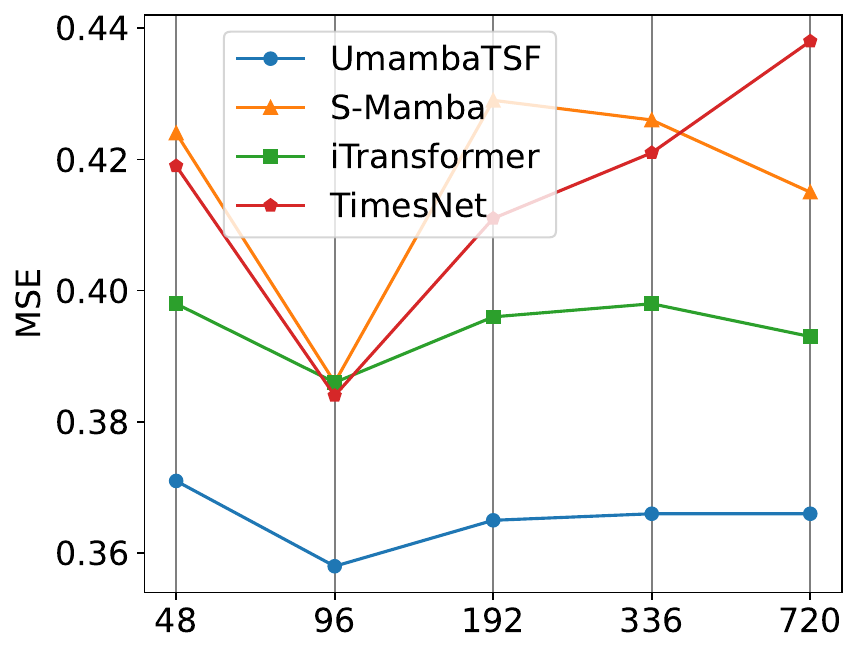}
    }
    \hspace{1mm}
    \subfloat[Weather]{
        \includegraphics[width=0.4\textwidth]{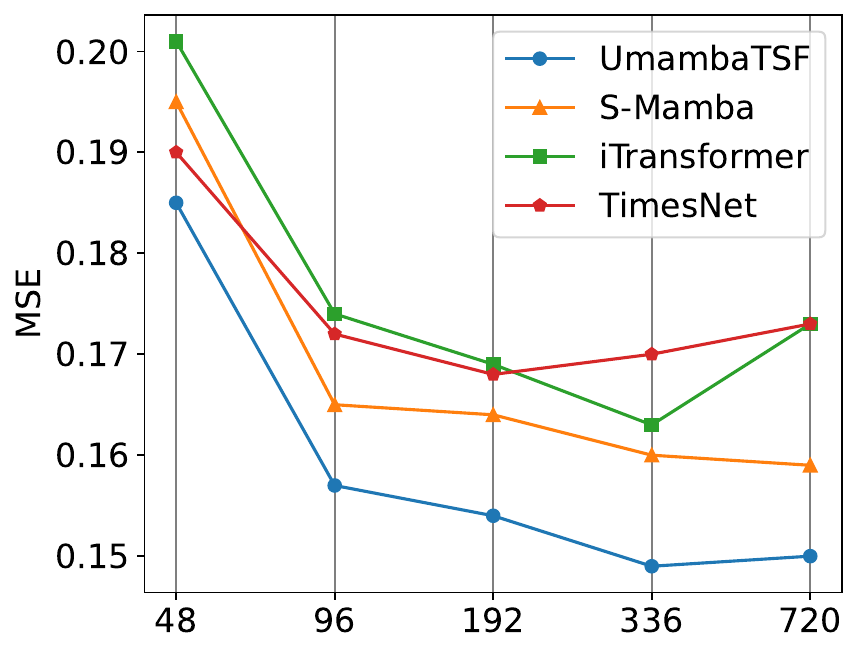}
    }
    \caption{Forecast performance with input length $L \in \{48, 96, 192, 336, 720\}$ and fixed prediction length $T = 96$. UmambaTSF demonstrates a significant advantage in MSE on the ETTh1 dataset, while on the Weather dataset, UmambaTSF gains more performance improvement from the extended lookback window length.}
    \label{fig:scale}
\end{figure}

Previous studies have shown that the forecasting performance of models like Transformers does not necessarily improve with an increasing lookback length due to the problem of distracted attention from the expanding input \cite{nie2023a,zeng2023transformers}. To validate whether our linear-complexity model, UmambaTSF, maintains its advantage across various lookback lengths, we conduct comparative analyses with three recent advanced models: S-Mamba, iTransformer, and TimesNet. These are tested on the ETTh1 and Weather datasets with lookback lengths set at 48, 96, 192, 336, and 720, and a forecasting window of 96. The results, depicted in Fig. \ref{fig:scale}, indicate that UmambaTSF consistently performs best on these datasets across all tested input windows. Notably, on the Weather dataset, as the lookback length increases, there is a corresponding enhancement in forecasting accuracy. This improvement is attributed not only to Mamba's ability to address long-range dependency issues but also to the MLP-based linear layers, which are better suited for leveraging larger historical datasets.

\subsubsection{Model Robustness}
\begin{figure}[h]
\centering
\includegraphics[width=0.45\textwidth]{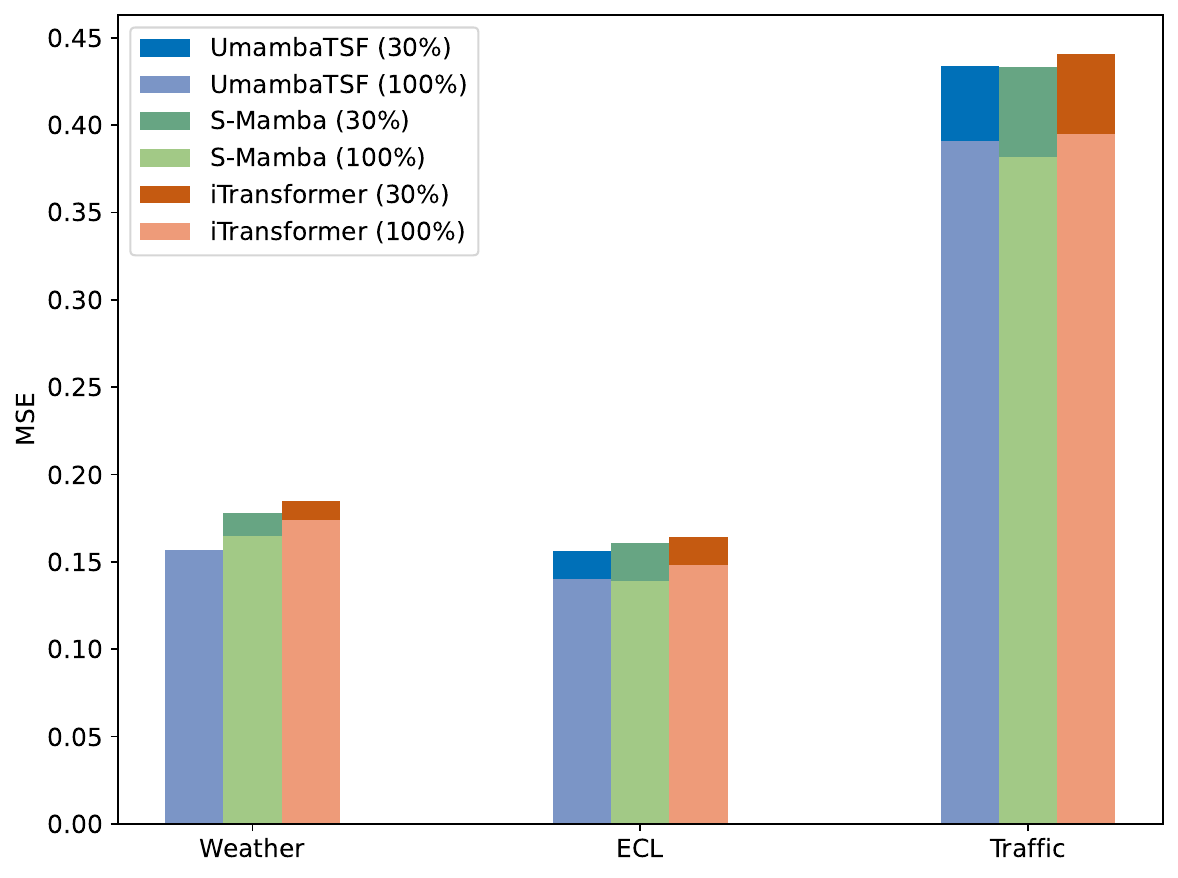}
\caption{Forecasting performance comparison among UMambaTSF, S-Mamba and iTransformer trained on 100\% samples with on
30\% samples. The lookback length $L = 96$ and the forecast length $T = 96$ for all datasets.}
\label{fig:robust}
\end{figure}

Time series forecasting is conducted on recorded sequential data, which typically does not allow for large-scale training sets. Solutions based on the Transformer architecture are often contested due to insufficient data volumes necessary for training robust models. To validate the robustness of our method, we conduct experiments on the Weather, ECL, and Traffic datasets to compare the performance of UmambaTSF, S-Mamba, and iTransformer across the full dataset and with 30\% of the original dataset. The results, depicted in Fig. \ref{fig:robust}, demonstrate that UmambaTSF maintains strong robustness, consistently outperforming or matching the advanced models iTransformer and S-Mamba, even when the dataset is reduced to only 30\%. Remarkably, on the Weather dataset, its performance using only 30\% of the data is identical to that achieved with the full dataset. This superior performance is likely attributed to UmambaTSF's powerful feature extraction capabilities and its streamlined, effective architecture. 

\section{CONCLUSION}
The goal of this paper is to achieve a win-win outcome in terms of prediction accuracy and computational cost for long-term time series forecasting using a linearly scalable approach. This goal has been realized by the proposed innovative method, UmambaTSF. The model employs a multi-scale feature extraction module, capable of capturing time series information across various scales. Integrating residual Mamba layers and a multi-scenario flexible transformation Mamba module, it significantly enhances both accuracy and versatility. 

In the experiments, we evaluate UmambaTSF on the 7 real-world datasets against 10 advanced baseline models. The results demonstrate that UmambaTSF achieves SOTA performance while maintaining computational costs low. Moreover, its compact and efficient design further enhances the model's generalization capability. This paper highlights Mamba's potential for time series forecasting from the perspectives of predictive performance, computational efficiency, and generalization ability. In the future, we plan to explore more channel processing techniques to further improve the accuracy of the forecasting model in noisy, multivariate environments.

%{\appendix[Proof of the Zonklar Equations]


%\begin{thebibliography}{1}
%\bibliographystyle{IEEEtran}


\vspace{1pt}

\begin{IEEEbiography}[{\includegraphics[width=1in,height=1.25in,clip,keepaspectratio]{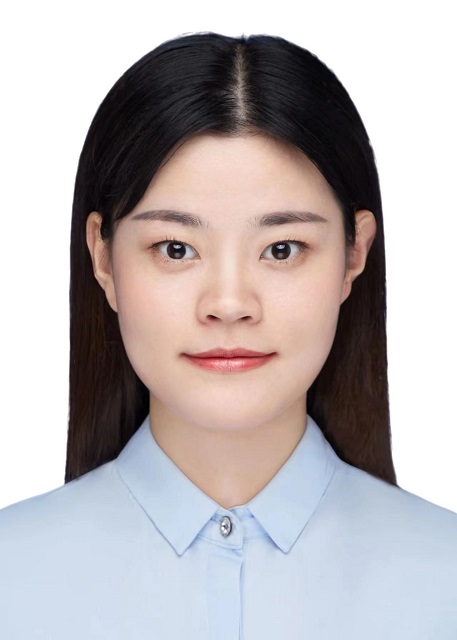}}]{Li Wu} received her M.Eng. degree from Tsinghua University in Beijing, China. She is currently pursuing her Ph.D. at Dalian University of Technology. Her primary research interests include time series modeling, spatiotemporal forecasting, and weather prediction. She is also a lecturer at Qinghai University and has published more than 10 papers, including in PPoPP and ICASSP. She is currently leading the National Natural Science Foundation project on ``Precipitation Forecasting Methods Based on Multi-Source Data Using Deep Learning and Its Application in the Sanjiangyuan Region'', and participated in several national key research and development programs.
\end{IEEEbiography}

\begin{IEEEbiography}[{\includegraphics[width=1in,height=1.4in]{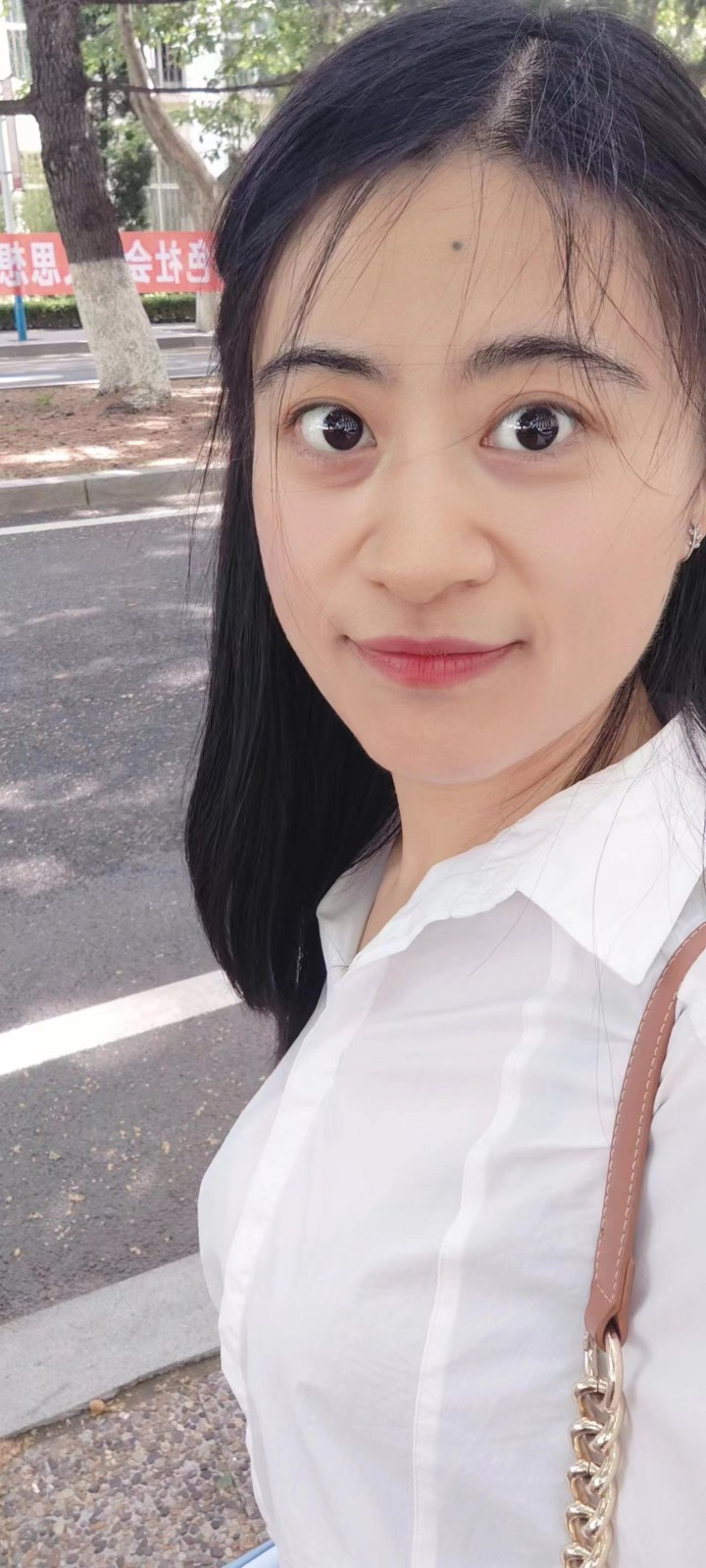}}]{Wenbin Pei} received her Ph.D. degree at Victoria University of Wellington, New Zealand, in 2021. She is currently an assistant professor at Dalian University of Technology. Her research interests include evolutionary computation and machine learning. She has published more than 30 papers in fully refereed international journals and conferences, including IEEE Transactions on Evolutionary Computation and AAAI. She was a program committee member for 6 international conferences and a co-chair of special sessions in 4 international conferences. She has been serving as reviewers for international journals, including IEEE Transactions on Evolutionary Computation, IEEE Transactions on Cybernetics, Evolutionary Computation journal (MIT press), IEEE Transactions on Emerging Topics in Computational Intelligence, Applied Soft Computing, and Artificial Intelligence in Medicine.
\end{IEEEbiography}
\begin{IEEEbiography}[{\includegraphics[width=1in,height=1.25in,clip,keepaspectratio]{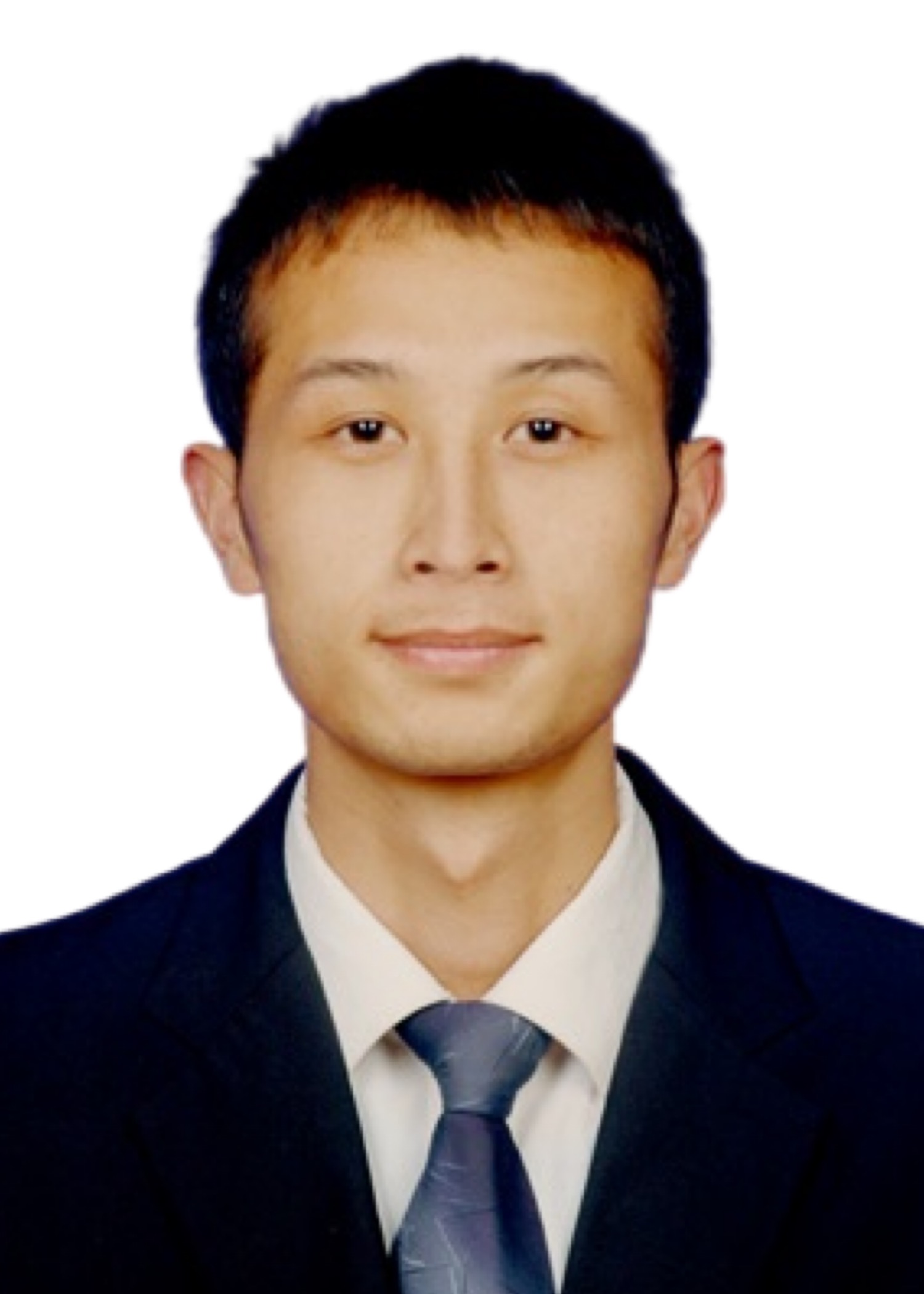}}]{JiuLong Jiao} received his B.Eng. and M.Eng. degrees from Harbin Institute of Technology. He is currently pursuing his Ph.D. at Dalian University of Technology. His primary research focuses on machine learning and multimodal few-shot learning.
\end{IEEEbiography}
\begin{IEEEbiography}[{\vspace{-4mm} \includegraphics[width=1in,height=1.4in]{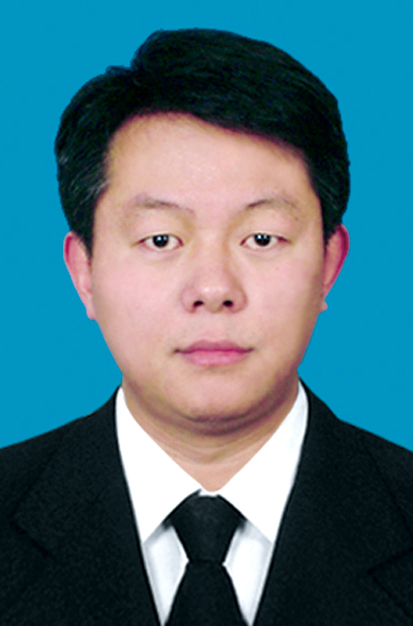}}]{Qiang Zhang} received the Ph.D. degree in Circuits and Systems from Xidian University, Xi'an, in 2002. He is currently a professor and the dean of the School of Computer Science and Technology, at Dalian University of Technology. His research interest includes bio-inspired computing and the related applications. Prof. Zhang has published more than 70 papers in fully refereed international journals and conferences. He was awarded National Science Fund for Distinguished Young Scholars in 2014, and was also selected as one of the state department special allowance experts. He has been serving as editorial board for 7 international journals and chairs of special issues in journals, such as Neurocomputing and International Journal of Computer Applications in Technology.
\end{IEEEbiography}

\vfill
\end{document}